\documentclass{ieeeaccess}

\usepackage{soul}
\usepackage{algorithm}
\usepackage{listings}
\usepackage{multirow}
\usepackage{lineno}
\usepackage{hyperref}
\usepackage{algpseudocode}
\usepackage{booktabs}
\usepackage{mathtools}
\soulregister\cite7
\soulregister\ref7
\soulregister\sc7
\usepackage{eurosym}
\usepackage{siunitx}
\usepackage{textgreek}
\usepackage{tablefootnote}
\renewcommand{\theenumi}{\Alph{enumi}}

\def\textsc#1{\textnormal{{\sc #1}}}
\usepackage{caption}
\DeclareCaptionFont{ieeeblue}{\color{accessblue}}
\DeclareCaptionLabelFormat{myformat}{\raggedright\figcapfont{\textbf{#1}\textbf{#2}}}
\def\BibTeX{{\rm B\kern-.05em{\sc i\kern-.025em b}\kern-.08em
T\kern-.1667em\lower.7ex\hbox{E}\kern-.125emX}}

\newcommand{\expnumber}[2]{{#1}\text{e}\textsuperscript{#2}}
\newcommand{\sctxt}[1]{{\sc #1}}
\newcommand{\red}[1]{{\color{red} #1}}
\soulregister{\sctxt}{7}
\soulregister{\red}{7}
\soulregister{\expnumber}{7}
\soulregister{\SI}{7}
\soulregister{\percent}{7}

\usepackage{blindtext}
\usepackage{hyperref}
\usepackage{nameref}

\newcounter{mylabelcounter}

\makeatletter
\newcommand{\labelText}[2]{%
\refstepcounter{mylabelcounter}%
\immediate\write\@auxout{%
 \string\newlabel{#2}{{\unexpanded{#1}}{\thepage}{{\unexpanded{#1}}}{mylabelcounter.\number\value{mylabelcounter}}{}}%
}
}
\makeatother

\makeatletter
\newcommand\footnoteref[1]{\protected@xdef\@thefnmark{\ref{#1}}\@footnotemark}
\makeatother

\begin{document}

\history{Received 16 October 2023, accepted 11 December 2023, date of publication 13 December 2023, date of current version 19 December 2023.}
\doi{10.1109/ACCESS.2023.3342472}

\title{Interpretable Classification of Wiki-Review Streams}

\author{\uppercase{Silvia García-Méndez}\authorrefmark{1}, \uppercase{Fátima Leal}\authorrefmark{2}, \uppercase{Benedita Malheiro}\authorrefmark{3}, and \uppercase{Juan Carlos Burguillo-Rial}\authorrefmark{1}}

\address[1]{Information Technologies Group, atlanTTic, University of Vigo, Vigo, Spain}

\address[2]{Research on Economics, Management and Information Technologies, Universidade Portucalense, Porto, Portugal}

\address[3]{ISEP, Polytechnic of Porto, Porto, Portugal \& Institute for Systems and Computer Engineering, Technology and Science, Porto, Portugal}

\tfootnote{This work was partially supported by Xunta de Galicia grant ED481B-2021-118, Spain; and Portuguese national funds through FCT –- Fundação para a Ciência e a Tecnologia (Portuguese Foundation for Science and Technology) -- as part of project UIDB/50014/2020.}

\markboth
{Silvia García-Méndez \headeretal: Interpretable Classification of Wiki-Review Streams}
{Silvia García-Méndez \headeretal: Interpretable Classification of Wiki-Review Streams}

\corresp{Corresponding author: Silvia García-Méndez (e-mail: sgarcia@gti.uvigo.es).}

\begin{abstract}
Wiki articles are created and maintained by a crowd of editors, producing a continuous stream of reviews. Reviews can take the form of additions, reverts, or both. This crowdsourcing model is exposed to manipulation since neither reviews nor editors are automatically screened and purged. To protect articles against vandalism or damage, the stream of reviews can be mined to classify reviews and profile editors in real-time. The goal of this work is to anticipate and explain which reviews to revert. This way, editors are informed why their edits will be reverted. The proposed method employs stream-based processing, updating the profiling and classification models on each incoming event. The profiling uses side and content-based features employing Natural Language Processing, and editor profiles are incrementally updated based on their reviews. Since the proposed method relies on self-explainable classification algorithms, it is possible to understand why a review has been classified as a revert or a non-revert. In addition, this work contributes an algorithm for generating synthetic data for class balancing, making the final classification fairer. The proposed online method was tested with a real data set from Wikivoyage, which was balanced through the aforementioned synthetic data generation. The results attained near-\SI{90}{\percent} values for all evaluation metrics (accuracy, precision, recall, and \textit{F}-measure).
\end{abstract}


\titlepgskip=-15pt

\maketitle

\section{Introduction}

Wiki-based platforms, like Wikipedia\footnote{Available at \url{https://en.wikipedia.org}, December 2023.}, WikiVoyage\footnote{Available at \url{https://www.wikivoyage.org}, December 2023.} or WikiNews\footnote{Available at \url{https://www.wikinews.org}, December 2023.} are collaboratively maintained by voluntary editors who share their wisdom about a topic, entity, or city. When editors create and refine wiki articles, they generate a continuous stream of events indistinctly referred to as edits or reviews. Specifically, wiki editors can add, edit, and revert reviews. As such, wikis are modern-day oracles maintained by and for the crowd, simultaneously empowering and impacting it. 

Moreover, this information-gathering model, known as crowdsourcing, accumulates the digital legacy of the crowd, allowing the scrutiny of interested parties. In this respect, wikis, discussion forums, blogs, and social networks can be mined to profile editors with the help of Artificial Intelligence (\textsc{ai}) \cite{Alattar2021,Jiang2023}. This activity is essential in crowdsourcing platforms since crowdsourced data are unmediated by default, exposing platforms to social manipulation. Examples of social manipulation include fake information in social networks, biased feedback in evaluation-based platforms, and undesired content in Wiki articles. Such adverse contributions may damage brands, products, services, and articles, affecting the overall reliability of the targeted platforms. 

One of the most popular techniques for spreading disinformation online is through standalone or coordinated brigades of false data generator bots. In the case of Wikipedia, sock-puppets -- individuals who create multiple online identities to increase their influence in online communities -- constitute a severe problem \cite{Zhang2021,Tajrian2023}. To mitigate misinformation, wikis rely on highly ranked editors -- administrators -- to patrol the contents. Interestingly, these threats to the reliability of wikis can only be offset by instantly reverting damaging reviews and swiftly outcasting unreliable editors \cite{Lageard:2021}. By classifying reviews and editors in real-time, the current work aims to address misinformation and reliability simultaneously. 

This work proposes an interpretable classification solution to recognize in real-time which reviews to revert. Therefore, this paper contributes with real-time transparent identification of deceitful wiki reviews and editors. By anticipating the reversion of undesirable reviews, this early discarding of reviews has a positive impact on both the quality and reliability of wiki data. The proposed method employs stream-based processing, updating the profiling and classification models on each incoming event. The profiling uses side and content-based features employing Natural Language Processing (\textsc{nlp}). Since the proposed method relies on self-explainable classification algorithms (\textit{e.g.}, decision trees), it is possible to understand why a review has been classified as a revert or a non-revert. 

In addition, this paper contributes with a data synthetic generation algorithm for class balancing, aiming to make the final classification fairer. In fact, synthetic data generation has been reported to be highly beneficial \cite{Mukherjee2021,Dankar2022} in (\textit{i}) testing stochastic and multispectral scenarios, (\textit{ii}) creating relevant scenarios absent in real data, (\textit{iii}) automatically labeling entries, (\textit{iv}) overcoming data restrictions and making the process more affordable, (\textit{v}) protecting sensible information, and (\textit{vi}) speeding up data analytic processes. However, it comes with relevant constraints that must be taken into account, such as (\textit{i}) the complexity for specific data scenarios, (\textit{ii}) bias and outliers that can be reflected from real data, (\textit{iii}) the dependent quality on the data source, and (\textit{iv}) laborious and time-consuming validation and evaluation against the original data. The proposed synthetic data generation module seeks to take advantage of the first three and the last benefits while aiming to address all four constraints pointed out in the literature.

The experiments were conducted with a real data set collected from Wikivoyage with \num{285698} reviews, including \num{8305} reverts, and \num{70260} editors. Despite the original imbalanced class distribution, the proposed method presents macro and micro class classification metrics near-\SI{90}{\percent}.

The rest of this paper is organized as follows.
Section~\ref{sec:2} overviews the relevant related work concerning profiling, classification, transparency, and fairness of wiki data and states the current contribution. Section~\ref{sec:3} introduces the proposed method, detailing the offline and stream processing. Section~\ref{sec:4} describes the experimental set-up and presents the empirical evaluation results considering the online revert classification and explanation. Finally, Section~\ref{sec:5} concludes and highlights the achievements and future work.

\section{Related work}
\label{sec:2}

In wiki-based platforms, problems such as transparency, fairness, and real-time modeling still need to be explored \cite {lealwiki2020}.

\subsection{Profiling}

Wiki profiling methods model editors through their interactions within the platform. In addition, in stream-based modeling, profiles are continuously updated and refined. Based on the contents of crowdsourced data, the literature contemplates multiple types of wiki profiling methods:

\begin{description}
 
 \item \textbf{Graph embedding} profiling is an unsupervised learning technique representing the learned graph nodes through low-dimensional vectors.  \cite{Heindorf:2019} created and represented profiles based on side features via graph embedding to detect unbiased vandalism. 
 
 \item \textbf{Stylometric} profiles are based on textual patterns of style, \textit{i.e.}, rely on the contents. \cite{Harpalani:2011} built stylometric profiles to detect vandalism in Wikipedia articles. \cite{Dauber:2017} used standard stylometric metrics (\textit{e.g.}, digit \textit{n}-gram frequency, word \textit{n}-gram frequency, \textit{etc.}) to identify the authorship in collaborative documents. \cite{Martinez-Rico:2019} identified style patterns using artificial neural networks to generate the linguistic model that represents a text.
 
 \item \textbf{Trust \& reputation} profiling represents the reliability of wiki editors. By definition, while trust is based on one-to-one relationships, reputation considers third-party experiences. Trust-based models are popular among wikis. \cite{Zhao:2011,Zhao:2013} proposed TrustWiki to establish the trustworthiness of wiki reviews based on the social context of editors. Hence, TrustWiki creates clusters of editors, using content and demographic features, and presents the reader with content from similar groups of editors. \cite{Adler:2012} implemented WikiTrust, which highlights trustworthy and untrustworthy words in wiki articles with different background colors. As such, WikiTrust explores content and side features. To prevent malicious and unreliable users, \cite{Kardan:2015} adopted SigmoRep to compute the reputation of editors in collaborative environments from side features. \cite{Paul:2015} used WikiTrust side and content-based features to recognize the authorship of crowdsourced content.
\end{description}

The literature shows several wiki editor and review profiling approaches that explore content, side, and social features. Besides, most surveyed works implemented offline processing. The only exception is the stream-based quality and popularity profiling proposed by \cite{lealwiki2019}, which enables model updating in real-time, along with the user profiling work of \cite{Garcia-Mendez2022} that identifies benign and malign human and non-human (bots) contributors. Since the classification problem in the latter work is different, the content of the review is not considered.

\subsection{Analysis of reviews}

According to the literature, the classification of wiki edits encompasses the detection of paid \cite{Joshi:2020}, puffery \cite{Bertsch:2021}, reverted \cite{Flock:2012,Segall2013,Kiesel:2017}, toxic \cite{Ibrahim:2018,Chakrabarty:2020} and vandal \cite{Heindorf:2019,Harpalani:2011,Martinez-Rico:2019,Zhao:2011,Paul:2015,Potthast:2008,Adler:2011, Javanmardi:2011,Mola-Velasco:2011,Alfonseca:2013,Tran:2013, Kumar:2015,Heindorf:2016,Shulhan:2016,Heindorf:2017,Sarabadani:2017, Liu:2019} reviews. Similarly, prediction focuses on review quality \cite{Sarkar:2019,Asthana:2021,Wong:2021} as well as on editor and article quality \cite{lealwiki2019,Ruprechter:2020,dang2016measuring,dang2017end, lewoniewski2017relative}.

\subsubsection{Vandalism detection}

Vandalism and error detection methods explore side and content-based features to identify unethical behaviors and unintentional errors. Specifically, unethical behaviors are practiced mainly by unregistered editors \cite{Alkharashi:2018}. In this context, the literature provides several vandalism detection methods employing distinct profiling and classification approaches.

Profiles use, separately or in combination, side and content-based features of the contributions. The Random Forest classifier \cite{Parmar2019} holds the best results and the highest popularity. Except for the detectors by \cite{Heindorf:2016,Heindorf:2017}, the remaining works implement offline processing. 

In addition to the features extracted by the described profiling methods, several vandalism detection solutions rely on scores from the Objective Revision Evaluation Service (\textsc{ores}), a public Application Programming Interface (\textsc{api}) for wiki platforms \cite{Halfaker:2020}. \textsc{ores} is a Machine Learning (\textsc{ml}) system that predicts the quality of edits and article drafts in real-time. Regarding edits, \textsc{ores} predicts the probability of being done in good faith, damaging, and reverted in the future. In the case of article drafts, \textsc{ores} returns the probability of being spam, vandalism, an attack, and \textsc{ok}. These scores are used as input features by many of the surveyed works, \textit{e.g.}, \cite{Heindorf:2019,Heindorf:2016,Heindorf:2017,Sarabadani:2017,Wong:2021}, to classify reviews. Moreover, \textsc{ores} is currently used on wiki platforms to help volunteers reduce the burden of manually screening content.

\subsubsection{Revert detection}
\label{sec:revert_dectection}

On wikis, reverting consists of completely removing a previous edit. Although it is a means to eliminate involuntary errors and malicious reviews, scant research is dedicated to revert classification: 

\begin{itemize}

 \item \cite{Flock:2012} designed a textual analysis algorithm to detect reverts and the corresponding target reviews. The algorithm analyses editor actions considering the inserted and deleted words. The unchanged paragraphs are removed, and the insert and delete actions are analyzed using text difference methods. The computational cost of this solution is the main drawback.
 
 \item \cite{Segall2013} proposed a content-agnostic, metadata-driven classification to detect Wikipedia reverts. The profiling model is based on the editor roles defined by Wikipedia and side-based information. The authors use a Support Vector Machine classifier.

\end{itemize}

All surveyed revert identification approaches implemented offline processing.

\subsubsection{Quality prediction}
\label{sec:quality}

Wikis, while unmediated collaborative environments, suffer from data quality and trustworthiness issues. To address this problem, predictive models can be used to anticipate the quality of individual wiki reviews, editors, and articles. To predict the quality of reviews, \cite{Sarkar:2019} employed orthographic similarity of lexical units to predict the quality of new reviews. Using side-based and stylometric profiles, the solution applies a deep neural network to extract quality indicators. \cite{Wong:2021} shared an annotated data set of English Wikipedia articles based on Wikipedia templates, \textit{e.g.}, original research, contradictory, unreliable sources, \textit{etc}. The data set was used to predict the content reliability using Logistic Regression \cite{Frey2023}, Random Forest, and Gradient Boosted Trees \cite{Mistry2021}. Except for the work by \cite{lealwiki2019}, all surveyed quality prediction works implemented offline techniques. 

\subsection{Transparency}

Interpretability and explainability are essential for users to understand \textsc{ml}-generated models and outputs, improving the experience and developing trust. While interpretability is defined as the ability to describe an \textsc{ml} model, explainability is the interface that enables the user to understand the model \cite{Adadi2018,Mollas2023}. \textsc{ml} models can be divided into interpretable and opaque. Opaque models behave as black boxes (\textit{e.g.}, artificial neural networks, or matrix factorization), whereas interpretable ones are self-explainable (\textit{e.g.}, decision trees, rules, or regression). A transparent model -- interpretable or explainable -- enhances decision-making and contributes to responsible \textsc{ml}.

Frequently, the \textsc{ml} algorithms used to mitigate the negative impact of malicious behaviors in wikis are opaque. To address this problem, several research works attempted to explain the operation of vandalism detectors \cite{Liu:2019,Subramanian:2019}, classifiers \cite{Mahajan:2021}, or profiling \cite{Sarker:2020,Klein:2021}. These explainability efforts take the form of the following:

\begin{description}
 
 \item \textbf{Graph-based} explanations rely on a knowledge representation graph to explain a context. \cite{Sarker:2020} and \cite{Klein:2021} represented and explained the relationship between wiki entities through graph-based profiling.
 
 \item \textbf{Model agnostic} explanations rely on surrogate models to explain the outcomes of opaque \textsc{ml} models. Model agnostic interpretability techniques include the Local Interpretable Model-agnostic Explanations (\textsc{lime}) \cite{Ribeiro:2016}, the Shapley Additive Explanations (\textsc{shap}) \cite{Lundberg:2017} and, more recently, the local trust-based explanation plugin (Explug) advanced by \cite{Leal:2022}. Considering wiki contents, \cite{Risch:2020} compared multiple explanation models, including \textsc{lime} and \textsc{shap}.
 
 \item \textbf{Word embedding} explanations are based on text processing, namely \textsc{nlp}. In this context, \cite{Mahajan:2021} combined word embedding with \textsc{lime} to assess the reliability of a toxic comment classifier while \cite{Qureshi:2019} integrated a word embedding technique with graph-based explanations.

 \item \textbf{Visual} explanations adopt non-textual formats easily interpretable by users. \cite{Liu:2019,Subramanian:2019} explained vandalism detection visually. While \cite{Liu:2019} correlated inputs and outputs with the parameter spaces based on the edit frequency and reverts, \cite{Subramanian:2019} analyzed statistically vandal behaviors and displayed the results graphically.

\end{description}

\subsection{Fairness}

Fairness embraces equal opportunity against biased algorithms or data to avoid prejudicial or unethical results. A fair model ensures that data biases do not affect its performance. While accuracy evaluates the performance of an \textsc{ml} model, fairness indicates the practical implications of deploying the model in the real world. Therefore, the selection of \textsc{ml} predictive algorithms must be guided not only by performance but also by fairness. 

Wikis tend to discriminate against unregistered or new editors regarding vandalism detection \cite{Alkharashi:2018,Ye:2021}, article ranking \cite{Lewandowski:2011} or reverts based on the contents of talk pages \cite{Ross:2014}. These issues have been addressed through class balancing, a pre-processing technique that balances the number of samples across classes. Regarding wiki vandalism detection, \cite{Potthast:2008} performed random over-sampling to address the imbalanced class problem, whereas \cite{Shulhan:2016} re-sampled wiki data with Local Neighbourhood Synthetic Minority Over-sampling Technique (\textsc{smote}).

\subsection{Research contribution}

The literature on real-time vandalism detection and revert classification reveals a lack of fundamental studies on vandal behavior \cite{Tramullas:2016}. Currently, misinformation and quality control in wiki platforms are addressed by a group of dedicated voluntary users named \textit{patrollers}\footnote{Available at \url{https://en.wikipedia.org/wiki/Wikipedia:Patrolling}, December 2023.} together with \textsc{ores}. When compared to \textsc{ores}, the proposed method classifies incoming edits and explains the verdict on a stream basis. Moreover, given its modular design, it enables the integration of the latest technological advances, such as using Large Language Models (\textsc{llm}s) to engineer new features for effective classification and better explainability. Such optional improvements in detection accuracy will lead to greater computational load.

Consequently, this work contributes to mitigating the vulnerabilities of the crowdsourcing model through real-time classification and explanation of the content posted in wikis. The designed pipeline, which applies existing \textsc{ai} methods, constitutes an original wiki vandalism detection method. Specifically, it employs:

\begin{itemize}
    \item Standard feature analysis, engineering, and selection techniques to ensure the high quality of the experimental data and take advantage of the full potential of the classification models.
    \item Online editor profiling to capture the expected profile evolution with time.
    \item Synthetic data generation techniques to create artificial samples and improve the fairness of the final classification.
    \item Stream-based \textsc{ml} classification and explainability to detect and justify which reviews to revert in real-time.
\end{itemize}

In summary, this work contributes with (\textit{i}) a stream-based method that, unlike existing solutions, analyses and exploits textual content for classification purposes; (\textit{ii}) generation of synthetic data to perform stochastic and multispectral tests; and (\textit{iii}) visual and natural language explanations of the classifications. Furthermore, the solution was validated with a comprehensive set of experiments, including ad hoc tests, to determine its performance in offline scenarios with different partition sets of the experimental data.

\section{Proposed method}
\label{sec:3}

\begin{figure*}[ht!]
\centering
\includegraphics[scale=0.13]{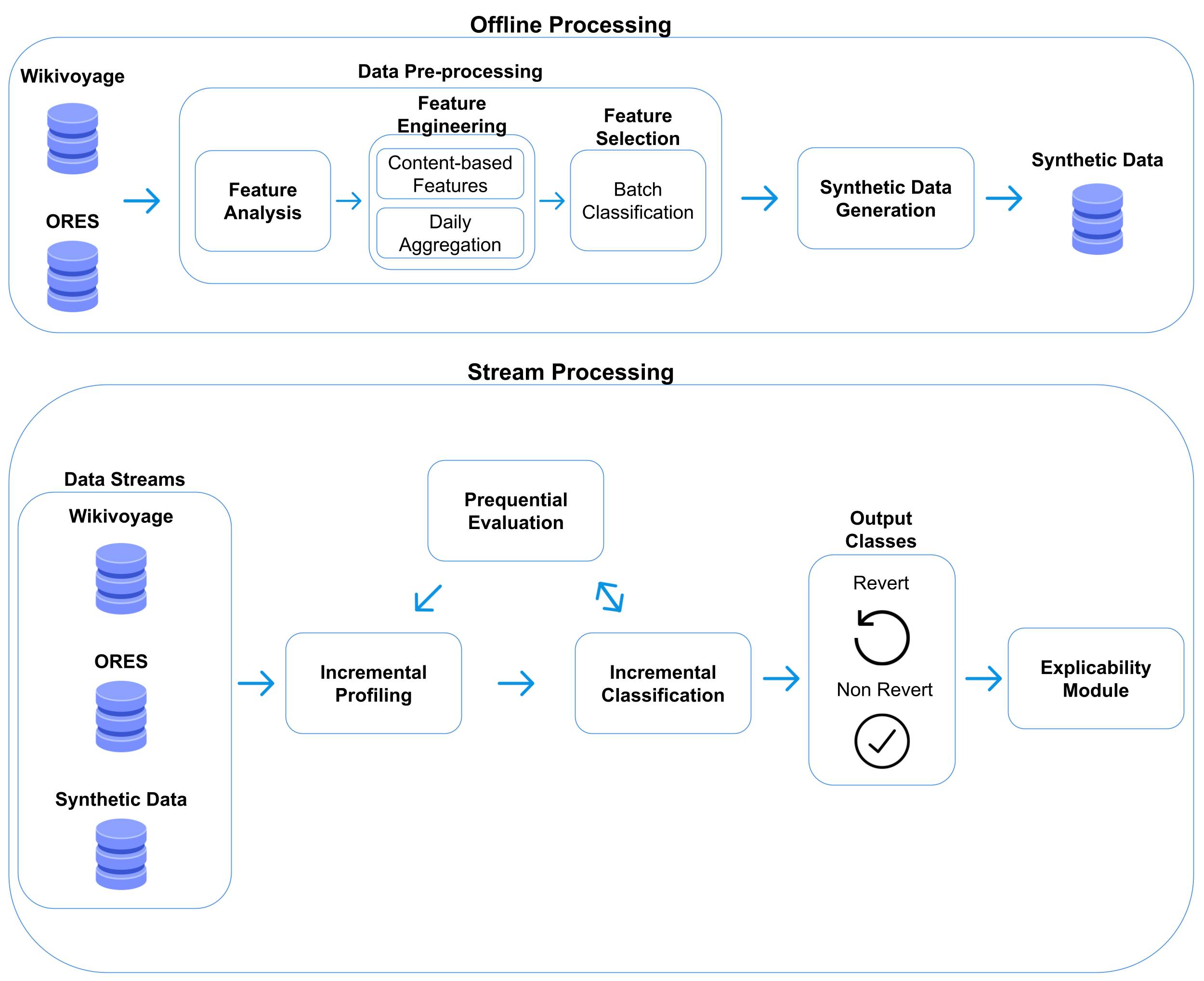}
\caption{\label{fig:method} Fair and transparent classification of Wikivoyage reviews as reverts and non-reverts.}
\end{figure*}

This paper proposes an explainable and fair method to identify which wiki reviews will be reverted. Figure \ref{fig:method} introduces the proposed multi-stage solution, which adopts offline synthetic wiki data generation for class balancing followed by stream-based classification of wiki reviews.
The offline stage encompasses (\textit{i}) data pre-processing (Section \ref{sec:data_processing}) based on feature-target pairwise correlation (Section \ref{sec:feature_analysis}), feature engineering (Section \ref{sec:feature_engineering}) and selection (Section \ref{sec:feature_selection}), and (\textit{ii}) synthetic data generation (Section \ref{sec:synthetic_data}) to balance the data set classes. The online stage performs (\textit{i}) incremental profiling (Section \ref{sec:profiling_stage}), (\textit{ii}) stream-based classification (Section \ref{sec:classification_stage}) evaluated through standard classification metrics, and (\textit{iii}) outcome explanation (Section \ref{sec:explanation_stage}) supported by model interpretability.

\subsection{Offline processing}
\label{sec:offline}

Offline processing comprises mainly data pre-processing techniques and synthetic data generation. However, the former is a three-phase stage. Algorithm \ref{alg:offline} overviews the offline processing method.

\begin{algorithm*}[!htbp]
 \scriptsize
 \caption{Offline processing algorithmic description.}\label{alg:offline}
 \begin{algorithmic}[0]
 \Function{offline\_processing}{$dataset$}

 \texttt{// Feature analysis}
 \State Spearman\_coefficients = compute\_Spearman(dataset);

 \texttt{// Feature engineering}
 \For{sample in dataset}
    \State sample.get\_text().process(); \texttt{// The specific text processing techniques will be detailed in Section \ref{sec:feature_eng_results}}
    \State sample.get\_text().compute\_features(); \texttt{// The specific features engineered will be detailed in Section \ref{sec:feature_eng_results}}
 \EndFor

 \texttt{// Feature selection}
 \State selected\_features = meta\_transformer\_wrapper(dataset, \textsc{rf}, configuration\_parameters);

 \texttt{// Synthetic data generation}
 \State synthetic\_data\_generation(); \texttt{// Detailed in Algorithm \ref{alg:building}.}

  \EndFunction
 \end{algorithmic}
\end{algorithm*}

\subsubsection{Data pre-processing}
\label{sec:data_processing}

Data pre-processing translates raw data into features usable by \textsc{ml} classifiers and selects the most promising independent features to predict the target feature. The new computed features relevant to the revert prediction task are employed to model editors and reviews. Specifically, data pre-processing is a three-phase stage composed of (\textit{i}) feature analysis, (\textit{ii}) feature engineering, and (\textit{iii}) feature selection tasks. First, feature analysis performs an in-depth screening of the features highly correlated with the target variable. Then, feature engineering enables valuable data generation for classification. Finally, feature selection takes the most relevant correlated features identified in feature analysis and those created during feature engineering. Moreover, the three data pre-processing techniques are applied offline before stream-based classification, as illustrated in Figure \ref{fig:method}.

\paragraph{Feature analysis}
\label{sec:feature_analysis}

The statistical dependence between the rankings of the independent features and the target feature is computed using Spearman's rank correlation coefficient \cite{DeWinter2016} as shown in Equation (\ref{spearman}), where $x$ and $y$ are the rank variables and $n$ represents the sample size.

\begin{equation}
r_s = \frac{n\sum{xy}-\sum{x}\sum{y}}{\sqrt{n\sum{x^2}-(\sum{x})^2}\sqrt{n\sum{y^2}-(\sum{y})^2}}
\label{spearman}
\end{equation}

This non-parametric measure of rank correlation assesses monotonic (linear or not) relationships among continuous and discrete features. Spearman correlation values range from -1 to +1, with the limits corresponding to the case when each feature is a perfect monotone function of the other.

\paragraph{Feature engineering}
\label{sec:feature_engineering}

This stage produces new side and content-derived features about articles, editors, and reviews to improve the classification of reviews as reverts and non-reverts. Side features characterize the properties of an entity (\textit{e.g.}, size of the revision or the number of links in the review). These features allow us to characterize the type of review performed by the editors (\textit{e.g.}, a large revision may indicate either a thorough correction, an excess of unnecessary changes, or the addition of spam/vandalism content). In contrast, content-derived features result from the analysis of the introduced or deleted text to provide \textsc{ml} models with in-depth knowledge of the content of the reviews.

Numeric features contain average values regarding editor revisions per article, revisions per week, articles revised per week, article quality probabilities from \textsc{ores}, review size, number of links, bad words, and number of inserted and deleted characters. Categorical features identify the creator of the revision, whether the editor is a bot, and hold the polarity of inserted and deleted text. In the end, textual features represent the cumulative characters and word $n$-grams of the inserted and deleted text. The generation of these features will be further explained in Section \ref{sec:feature_eng_results}.

Finally, the original data set consisting of individual timestamped reviews and related features is transformed into daily reviews and associated features per editor. The remaining stages explore these editor daily activity features instead of the original individual features.

\paragraph{Feature selection}
\label{sec:feature_selection}

Feature selection is performed through a meta-transformer wrapper method. Mainly, a meta-transformer method can be used with any estimator for feature selection, while a wrapper method allows the exploitation of an underlying \textsc{ml} model for the feature importance computation \cite{Garcia-Mendez2022}. It wraps the classification algorithm -- Random Forest (\textsc{rf}) classifier -- and selects features based on importance weights. The algorithm establishes relative feature importance using a forest of trees to find meaningful features and, thus, reduce the feature space dimension. The features are considered irrelevant if the corresponding importance of the feature values is below the specified threshold. The resulting profile feature vector comprises:

\begin{itemize}
 \item Side features related to editors;
 \item \textsc{ores} probabilities related to articles and reviews;
 \item Side and content-derived features related to reviews.
\end{itemize}

\subsubsection{Synthetic data generation}
\label{sec:synthetic_data}

Synthetic data generation allows testing \textsc{ml} models in a fully stochastic and multispectral scenario, including significant layouts absent from real data. The proposed synthetic data generation module is mainly used to balance the experimental data set.

More in detail, the designed data generation method produces feasible incremental samples of the editor's daily activity concerning the revert category. This process allows us to balance the experimental data set. Note that only reverted entries are created since they are less represented. The created artificial samples include the features listed in Table \ref{tab:daily_features}, except for the char and word $n$-grams. The latter are randomly selected by using a clustering procedure. To maintain the inter-feature correlation (see Section \ref{sec:syntheticdata}), the data generated for each feature is based on its statistical measures (quartile distribution, median, minimum and maximum values) considering four intervals: (\textit{i}) from minimum to the first quartile (Q1); (\textit{ii}) from Q1 to median; (\textit{iii}) from median to third quartile (Q3); (\textit{iv}) from Q3 to the maximum. Algorithm \ref{alg:building} summarizes the synthetic data generation procedure.

\begin{algorithm*}[!htbp]
 \scriptsize
 \caption{It creates $count$ revert samples using cluster-based filtering and maintaining inter-feature correlation.}\label{alg:building}
 \begin{algorithmic}[0]
 \Function{synthetic\_data\_generation}{$min$, $Q1$, $median$, $Q3$, $max$, $count$}
 
 \State $ranges = \{min, Q1, median, Q3, max\}$; \texttt{// Quartile distribution}
 
 \State $synthetic\_data = []$; \texttt{// Synthetic samples}
 
 \For{$r \in ranges-1$}
 
 \For{$i \in count/4$}
 
 \State $synthetic\_entry = []$; \texttt{// Synthetic sample}
 
 \For{$f \in features$} \texttt{// Table \ref{tab:daily_features} features}
 
 \State $Q1 = \textit{K}means[f,r,r+1]$
 
 \State $Q3 = \textit{K}means[f,r,r+1]$
 
 \State \texttt{//Random sample from Q1 to Q3 obtained by the \textit{K}-means}
 
 \State $synthetic\_entry[f]=random(Q1,Q3)$
 
 \EndFor
 
 \State $synthetic\_data.append(synthetic\_entry)$
 
 \EndFor
 
 \EndFor
 
 \State $synthetic\_data.sortByTimestamp()$ \texttt{// Returns the synthetic samples}
 
  \EndFunction
 \end{algorithmic}
\end{algorithm*}

For example, to generate a Q1 value for numeric feature $f$, it retrieves all samples from the original data set with an $f$ value between its Q1 and median. Then, it generates the remaining features based on the above subset with random values between Q1 (minimum) and Q3 (maximum). The latter avoids the generation of outliers (caused by applying cluster-based filtering and having a minimal subset) and enhances the dispersion of the synthetically generated data. Once all non-cumulative numeric features of the synthetic feature vector are generated, the values of cumulative features are randomly selected from a hash map that holds all possible values for those features in the subset. The date is randomly selected within the period of the experimental data set to prevent revert and non-revert samples from grouping and, thus, obtaining an unrealistic time distribution.

\subsection{Stream processing}

Stream processing is a three-phase stage that involves incremental profiling, classification, and the generation of explainable descriptions regarding the predictions. Algorithm \ref{alg:stream} overviews this process.

\begin{algorithm*}[!htbp]
 \scriptsize
 \caption{Stream processing algorithmic description.}\label{alg:stream}
 \begin{algorithmic}[0]
 \Function{stream\_processing}{$dataset$}

 \texttt{// Incremental profiling}
 \State dataset = compute\_average\_values(dataset);
 \State dataset = compute\_cumulative\_values(dataset);
 
 \texttt{// Incremental classification}
 \State ml\_models = load\_ml\_models();
 \For{model in ml\_models}
    \For{sample in dataset}
        \State predict(model, sample);
        \State evaluate(model, sample);
        \State train(model, sample);
        \State print(model.evaluation\_metrics());
    \EndFor
 \EndFor

 \texttt{// Explainability}
 \State explainability\_graph = \textsc{rf}.compute\_graph(dataset.get\_random\_sample());
 \State visualize(explainability\_graph);
 \State nl\_description = apply\_template(dataset.get\_random\_sample());
 \State print(nl\_description);

  \EndFunction
 \end{algorithmic}
\end{algorithm*}

\subsubsection{Incremental profiling}
\label{sec:profiling_stage}

The incremental profiling models the online evolution of the editor's daily activity through the selected feature vectors from the offline data pre-processing stage. Editor profiles are incrementally updated using the balanced data set as a data stream. The built profiles encompass side features (\textit{e.g.}, number of inserted or deleted characters) and content-derived ones (\textit{e.g.}, the polarity, bad words, or $n$-grams of reviews). Several numeric features store cumulative averages and sums. 

\subsubsection{Incremental classification}
\label{sec:classification_stage}

This work comprises both batch and stream-based \textsc{ml} classification. The batch experiments select the most promising features and classification algorithm (baseline results), whereas the stream-based experiment explores the batch findings online. 
The binary classification algorithms selected are well-known interpretable models with promising performance \cite{Heindorf:2019,Mi2020,Haffar2022,Hossin2023}:

\begin{itemize}
 \item Naive Bayes (\textsc{nb}) is a simple probabilistic classifier based on Bayes' theorem \cite{Berrar2019NB}.
 \item Ridge Classifier (\textsc{rc}) exploits Ridge regression by converting the target feature values into \{-1, 1\} \cite{sarker2021machine}.
 \item Decision Tree (\textsc{dt}) is a discrete-target predictive model based on traversing a tree structure from observation branches (conjunctions of features) to a target class label leaf \cite{Trabelsi2019}.
 \item Random Forest (\textsc{rf}) is an ensemble learning model based on multiple \textsc{dt} classifiers \cite{Parmar2019}.
 \item Boosting Classifier (\textsc{bc}) is an ensemble of weak predictive models that allows the optimization of a differentiable loss function \cite{Bentejac2021}.
\end{itemize}

The model evaluation relies on standard metrics: classification accuracy, precision, recall, and \textit{F}-measure in macro and micro-averaging computing scenarios. Furthermore, macro results enable comprehensive evaluation considering the whole set of target classes, while micro-averaging considers the target classes individually. The latter is beneficial in imbalanced classification problems \cite{Okkalioglu2022,Hafeez2023,Vanacore2023}. Finally, run-time is measured to compare the performance of the different models.

\subsubsection{Explainability}
\label{sec:explanation_stage}

The proposed method relies on interpretable classifiers to explain classification outcomes. Being self-explainable, interpretable models can make their reasoning explicit, offering insight into the classification process. Decision rules, decision trees, Naive Bayes, and logistic regression are examples of interpretable binary classification algorithms.

In the current case, the selected classification algorithms are interpretable and, thus, can explain why a review has been classified as a revert or a non-revert. This explainability may rely on graph-based, natural language, visual, or hybrid formats to present the user with the reasons learned by the classifier, \textit{e.g.}, the relevant \textsc{dt} path, decision rules, or the impact of the different features on the outcome.

\section{Experimental results}
\label{sec:4}

All experiments were performed using a server with the following hardware specifications:
\begin{itemize}
 \item Operating System: Ubuntu 18.04.2 LTS 64 bits
 \item Processor: Intel\@Core i9-9900K \SI{3.60}{\giga\hertz}
 \item RAM: \SI{32}{\giga\byte} DDR4 
 \item Disk: \SI{500}{\giga\byte} (7200 rpm SATA) + \SI{256}{\giga\byte} SSD
\end{itemize}

The experiments comprise (\textit{i}) offline feature analysis, engineering, and selection, (\textit{ii}) offline synthetic data generation for class balancing, (\textit{iii}) incremental profiling, online classification with a balanced data stream and prediction explanation.

\subsection{\label{tuning_training_results} Data set}

The data collection\footnote{Data and code will be available from the corresponding author on reasonable request.} relied on a well-known set of utilities for extracting and processing MediaWiki\footnote{Available at \url{https://pypi.org/project/mediawiki-utilities}, December 2023.} data in Python. To retrieve the contents of the reviews, direct \textsc{get} requests were issued to the Wikivoyage endpoint\footnote{Available at \url{https://en.wikivoyage.org/w/api.php}, December 2023.} using the \textit{compare} action functionality. The data span from 1\textsuperscript{st} January 2004 to 31\textsuperscript{st} December 2019 contains \num{285698} samples from \num{70260} editors regarding \num{3369} different articles. Considering the target feature, the data is deeply imbalanced with a total of \num{8305} reverted reviews (\SI{0.03}{\percent} of the samples).

\begin{table}[!htbp]
\scriptsize
\centering
\caption{\label{tab:datasets} Wikivoyage data set transformation.}
 \begin{tabular}{llS[table-format=6]} 
 \toprule
 \bf Data set & \bf Contents & {\bf Total} \\
 \midrule
 \multirow{5}{*}{\textbf{Original} (imbalanced)} & Articles & 3369\\ 
 & Editors & 70260\\ 
 & Reviews / editor: & 285698\\
 & ~~~~~~Non-reverts & 277393\\
 & ~~~~~~Reverts & 8305\\
 \midrule
 \multirow{5}{*}{\textbf{Daily original}\tablefootnote{It groups editor activity per day.} (imbalanced)} & Articles &
 3369\\ 
 & Editors & 70260\\ 
 & Daily reviews / editor: & 45353\\
 & ~~~~~~Non-reverts & 41996\\
 & ~~~~~~Reverts & 3357\\
 \midrule
 \multirow{5}{*}{\textbf{Daily synthetic}\tablefootnote{It creates new synthetic samples per editor and day (only for imbalanced revert activity).} (imbalanced)} & Articles & 3369\\ 
 & Editors & 70260\\ 
 & Daily reviews / editor: & 40000\\
 & ~~~~~~Non-reverts & 0\\
 & ~~~~~~Reverts & 40000\\
 \midrule
 \multirow{5}{*}{\textbf{Daily combined} (balanced)} & Articles & 3369\\ 
 & Editors & 70260\\ 
 & Daily reviews / editor: & 85353\\
 & ~~~~~~Non-reverts & 41996\\
 & ~~~~~~Reverts & 43357\\
 \bottomrule
 \end{tabular}
\end{table}

\subsection{Offline processing}
\label{sec:offline_results}

As previously mentioned, offline processing comprises several relevant tasks: (\textit{i}) offline feature analysis, engineering and selection; and (\textit{ii}) offline synthetic data generation for class balancing.

\subsubsection{\label{tuning_training} Data pre-processing}

The data pre-processing starts with the statistical dependence analysis described in Section \ref{sec:feature_analysis} over the rankings of the numeric features listed in Table \ref{tab:features}. These exclude identifier features \numrange{1}{5} and textual features 11 and 12. The correlation results were computed using Spearman's rank correlation coefficient. Finally, it performs feature engineering and selection.

\paragraph{Feature analysis}

Table \ref{tab:features} presents the independent (\numrange{1}{22}) and target (23) features considered for the classification of reviews as reverts and non-reverts. The correlation between the independent and the target features in Table \ref{tab:features} is moderate and can be grouped into:

\begin{itemize}
 \item \textbf{Features with negative correlation}: from 6 to 8, 14, 19 (false damaging \& true good faith probabilities), 20 (\textsc{e} probability), 21 (\textsc{ok} probability), 22 (star \& stub probabilities).

 \item \textbf{Features with positive correlation}: 9, 10, 13, from 15 to 19 (true damaging \& false good faith probabilities), 20 (except \textsc{e} probability), 21 (except \textsc{ok} probability), 22 (except star \& stub probabilities).
\end{itemize}

\begin{table}[!htbp]
\scriptsize
\centering
\caption{\label{tab:features} Features considered for the classification and target feature.}
\begin{tabular}{cl} 
\toprule
\bf Number & \bf Feature name \\\midrule
1 & Date \\
2 & Review \textsc{id} \\
3 & Editor name and \textsc{id} \\
4 & Creator name and \textsc{id} \\
5 & Article name \\
6 & Bot flag \\
7 & Editor is the creator of the article \\
8 & Size of the revision \\
9 & Number of links in the review \\ 
10 & Number of repeated links in the review \\ 
11 & Inserted characters \\ 
12 & Deleted characters \\
13 & Amount of inserted characters \\ 
14 & Amount of deleted characters \\
15 & Amount of common reverted words added \\ 
16 & Amount of bad words added \\
17 & Polarity of inserted characters \\ 
18 & Polarity of deleted characters \\
19 & \textsc{ores} edit quality probability: false/true damaging \& good faith \\ 
20 & \textsc{ores} item quality probability: \textsc{a/b/c/d/e} \\
21 & \textsc{ores} article quality probability: \textsc{ok}/attack/spam/vandalism \\ 
22 & \textsc{ores} article quality probability (wp10): \textsc{b/c/fa/ga}/start/stub \\
\cmidrule[0.01em](lr){1-2}
23 & Revert flag \\
\bottomrule
\end{tabular}
\end{table}

\paragraph{Feature engineering}
\label{sec:feature_eng_results}

\begin{sloppypar}
The contents of the revisions are processed with the English Natural Language Processing pipeline optimized for \textsc{cpu}, named \texttt{\small en\_core\_web\_lg}\footnote{Available at \url{https://spacy.io/models/en\#en_core_web_lg}, December 2023.} provided by the spaCy library\footnote{Available at \url{https://spacy.io}, December 2023.}.
This processing removes \textsc{url} instances and special characters like accents. Then, it lemmatizes the resulting text and removes stop words\footnote{Available at \url{https://gist.github.com/sebleier/554280}, December 2023.}. The polarity of the revisions, considering added and deleted characters, is computed with the \texttt{\small spaCyTextBlob}\footnote{Available at \url{https://spacy.io/universe/project/spacy-textblob}, December 2023.} pipeline that performs sentiment analysis using the \texttt{\small TextBlob} library\footnote{Available at \url{https://github.com/sloria/TextBlob}, December 2023.}. To determine the number of common words reverted and bad words in revisions, it reuses the corresponding lists of words provided by Wikimedia Meta-wiki\footnote{Available at \url{https://meta.wikimedia.org/wiki/Research:Revision_scoring_as_a_service/Word_lists/en}, December 2023.}. The char and word $n$-grams are extracted from the accumulated textual data (see Section \ref{sec:profiling_stage}) with the help of the \texttt{\small CountVectorizer}\footnote{Available at \url{https://scikit-learn.org/stable/modules/generated/sklearn.feature_extraction.text.CountVectorizer.html}, December 2023.} Python library. Based on performance tests, the configuration parameters were set to \texttt{\small max\_df\_in=0.7}, \texttt{\small min\_df\_in=0.001}, \texttt{\small wordgram\_range\_in=(1,4)}, \texttt{\small chargram\_range\_in=(1,4)}, \texttt{\small max\_features\_in=None}.

Finally, it aggregates the individual reviews and associated features into daily reviews and associated features per editor, removing the hour and minute from the date. The remaining stages explore these daily features.
\end{sloppypar}

\paragraph{Feature selection}

The \texttt{\small SelectFromModel}\footnote{Available at \url{https://scikit-learn.org/stable/modules/generated/sklearn.feature_selection.SelectFromModel.html}, December 2023.} feature selection algorithm wraps the \textsc{rf} classifier to identify the higher-importance features. The configuration parameters were set to \texttt{\small n\_estimators=500}, \texttt{\small n\_jobs=-1}, \texttt{\small random\_state=0}. 
These experiments use a reduced balanced subset comprising \num{3357} revert and \num{4000} non-revert samples and 10-fold cross-validation \cite{Berrar2019} to avoid over-fitting, biased, or over-estimated values. Table \ref{tab:daily_features} lists the independent features selected for classifying reviews as reverts and non-reverts (features from 1 to 18 correspond to side features and features 19 and 20 to content features). To establish the best set of features, the batch classifier explored the three sets identified in Table \ref{tab:daily_features}:

\begin{enumerate}
 \item Side features related to editors;
 \item \textsc{ores} probabilities related to articles and reviews plus A features; 
 \item Content-derived features related to reviews plus B features.
\end{enumerate}

Ultimately, only some char and word $n$-grams features were discarded.

\begin{table}[!htbp]
\scriptsize
\centering
\caption{\label{tab:daily_features} Independent features selected for the classification.}
\begin{tabular}{c|c|c|cl} 
\toprule

\multicolumn{1}{c}{} & \multicolumn{1}{c}{} & \multicolumn{1}{c}{} & \bf Number & \bf Feature name \\
\midrule
\multirow{20}{*}{\rotatebox[origin=c]{90}{\bf Set C}} & \multirow{9}{*}{\rotatebox[origin=c]{90}{\bf Set B}} & \multirow{5}{*}{\rotatebox[origin=c]{90}{\bf Set A}} & 1 & Bot flag \\
& & & 2 & Editor is the creator of the article \\
& & & 3 & Average number of revisions per article \\
& & & 4 & Average number of revisions per week \\
& & & 5 & Average number of articles revised per week \\ 

& & \multicolumn{1}{c}{} & 6 & Average \textsc{ores} edit quality probability \\
& & \multicolumn{1}{c}{} & 7 & Average \textsc{ores} item quality probability \\
& & \multicolumn{1}{c}{} & 8 & Average \textsc{ores} article quality probability \\
& & \multicolumn{1}{c}{} & 9 & Average \textsc{ores} article quality probability (wp10) \\ 

& \multicolumn{1}{c}{} & \multicolumn{1}{c}{} & 10 & Average size of the revision \\
& \multicolumn{1}{c}{} & \multicolumn{1}{c}{} & 11 & Average number of links in the revision \\
& \multicolumn{1}{c}{} & \multicolumn{1}{c}{} & 12 & Average number of repeated links in the revision \\
& \multicolumn{1}{c}{} & \multicolumn{1}{c}{} & 13 & Average number of common reverted words \\
& \multicolumn{1}{c}{} & \multicolumn{1}{c}{} & 14 & Average number of bad words \\
& \multicolumn{1}{c}{} & \multicolumn{1}{c}{} & 15 & Average number of inserted characters \\
& \multicolumn{1}{c}{} & \multicolumn{1}{c}{} & 16 & Average number of deleted characters \\
& \multicolumn{1}{c}{} & \multicolumn{1}{c}{} & 17 & Average polarity of inserted characters \\
& \multicolumn{1}{c}{} & \multicolumn{1}{c}{} & 18 & Average polarity of deleted characters \\
& \multicolumn{1}{c}{} & \multicolumn{1}{c}{} & 19 & Cumulative char and word $n$-grams of inserted text \\
& \multicolumn{1}{c}{} & \multicolumn{1}{c}{}& 20 & Cumulative char and word $n$-grams of deleted text \\
\bottomrule
\end{tabular}
\end{table}

Finally, Table \ref{tab:batch_results} lists the results of the four offline \textsc{ml} classifiers with these sets of features using 10-fold cross-validation. The non-revert and revert classes correspond to \#0 and \#1, respectively. Set \textsc{a} of features report near 60 \% - 70 \% accuracy, precision, macro recall, and macro and non-revert \textit{F}-measure values. Even though non-revert values for recall are promising for the \textsc{nb} and \textsc{rc} classifiers, revert values for recall and \textit{F}-measure are significantly low. The results with set \textsc{b} (comprising set \textsc{a} and \textsc{ores} probabilities) are generally better for all classifiers and metrics. Results improve further with set \textsc{c}, attaining near 80 \% with \textsc{rf} and \textsc{bc} in all metrics. Set C comprises side features related to editors, \textsc{ores} features related to articles, plus content-derived features related to reviews. The best classifier considering all metrics is \textsc{rf}.

\begin{table*}[htb]
\scriptsize
\centering
\caption{\label{tab:batch_results}Offline revert classification results with a balanced data set of 7357 original samples using 90 \% for training and 10 \% for test (10-fold cross-validation).}
\begin{tabular}{ccccccccccccS[table-format=3.2]}
\toprule
\bf {Set} & \bf {Classifier} & \bf {Accuracy} & \multicolumn{3}{c}{\bf Precision} & \multicolumn{3}{c}{\bf Recall} & \multicolumn{3}{c}{\bf \textit{F}-measure} & {\bf Time}\\
\cmidrule(lr){4-6}
\cmidrule(lr){7-9}
\cmidrule(lr){10-12}
 & & & Macro & \#0 & \#1 & Macro & \#0 & \#1 & Macro & \#0 & \#1 & {(s)}\\
\midrule
\multirow{5}{*}{A} 
& \textsc{nb} & 0.60 & 0.67 & 0.58 & 0.77 & 0.56 & 0.96 & 0.17 & 0.50 & 0.72 & 0.27 & 0.02 \\
& \textsc{rc} & 0.61 & 0.69 & 0.59 & 0.78 & 0.58 & 0.95 & 0.20 & 0.53 & 0.73 & 0.32 & 0.04 \\
& \textsc{dt} & 0.58 & 0.57 & 0.61 & 0.54 & 0.57 & 0.63 & 0.51 & 0.57 & 0.62 & 0.52 & 0.15 \\
& \textsc{rf} & 0.59 & 0.59 & 0.63 & 0.56 & 0.59 & 0.63 & 0.55 & 0.59 & 0.63 & 0.55 & 2.09 \\
& \textsc{bc} & 0.64 & 0.64 & 0.64 & 0.64 & 0.62 & 0.77 & 0.47 & 0.62 & 0.70 & 0.54 & 3.51 \\
\midrule

\multirow{5}{*}{B} 
& \textsc{nb} & 0.65 & 0.69 & 0.62 & 0.77 & 0.62 & 0.92 & 0.33 & 0.60 & 0.74 & 0.46 & 0.02 \\
& \textsc{rc} & 0.65 & 0.67 & 0.63 & 0.71 & 0.63 & 0.86 & 0.40 & 0.62 & 0.73 & 0.51 & 0.05 \\
& \textsc{dt} & 0.67 & 0.67 & 0.72 & 0.62 & 0.67 & 0.64 & 0.70 & 0.67 & 0.68 & 0.66 & 0.68 \\
& \textsc{rf} & 0.74 & 0.73 & 0.76 & 0.71 & 0.74 & 0.75 & 0.72 & 0.73 & 0.75 & 0.71 & 2.54 \\
& \textsc{bc} & 0.69 & 0.69 & 0.69 & 0.68 & 0.68 & 0.77 & 0.58 & 0.68 & 0.73 & 0.63 & 16.65 \\
\midrule

\multirow{5}{*}{\bf C} 
& \textsc{nb} & 0.59 & 0.72 & 0.57 & 0.86 & 0.55 & 0.98 & 0.11 & 0.46 & 0.72 & 0.20 & 0.90 \\
& \textsc{rc} & 0.76 & 0.77 & 0.73 & 0.80 & 0.75 & 0.87 & 0.62 & 0.75 & 0.79 & 0.70 & 2.52 \\
& \textsc{dt} & 0.73 & 0.74 & 0.78 & 0.69 & 0.74 & 0.71 & 0.77 & 0.73 & 0.74 & 0.73 & 14.09 \\
& \textsc{rf} & \bf 0.83 & \bf 0.84 & \bf 0.82 & \bf 0.85 & \bf 0.83 & \bf 0.88 & \bf 0.77 & \bf 0.83 & \bf 0.85 & \bf 0.81 & 7.58 \\
& \textsc{bc} & 0.83 & 0.83 & 0.82 & 0.84 & \bf 0.83 & 0.88 & 0.77 & 0.83 & 0.85 & 0.81 & 307.20 \\
\bottomrule
\end{tabular}
\end{table*}

\subsubsection{Synthetic data generation}
\label{sec:syntheticdata}

The quality of the synthetic data was determined by statistically comparing the synthetic against the original daily data. Table \ref{tab:comparative_original_synthetic} displays the results, excluding the features without statistical variation (1, 2, 14 in Table \ref{tab:daily_features}). 

\begin{table*}[htb]
\centering
\scriptsize
\caption{Relative change (\%) between the synthetic and original daily data related to first, second, and third quartiles.}\label{tab:comparative_original_synthetic}
\begin{tabular}{l S[table-format=2.4] S[table-format=2.4] S[table-format=2.4] S[table-format=2.4] S[table-format=2.4] S[table-format=2.4] S[table-format=2.4]}
\toprule
&{\bf 3} &{\bf 4} &{\bf 5} & \multicolumn{4}{c}{\bf 6} \\
\cmidrule(lr){5-8}
& & & & \multicolumn{1}{c}{damaging false} & \multicolumn{1}{c}{damaging true} & \multicolumn{1}{c}{goodfaith false} & \multicolumn{1}{c}{goodfaith true} \\
\midrule
Q1 &18.61 &19.33 &24.81 &2.74 &3.21 &0.12 &0.15 \\
Q2 &20.63 &11.41 &9.51 &0.01 &0.06 &0.06 &0.06 \\
Q3 &17.99 &20.11 &13.44 &3.26 &2.80 &0.15 &0.12 \\
\bottomrule
& \multicolumn{5}{c}{\bf 7} & \multicolumn{2}{c}{\bf 8} \\
\cmidrule(lr){2-6}
\cmidrule(lr){7-8}
&\multicolumn{1}{c}{A} &\multicolumn{1}{c}{B} & \multicolumn{1}{c}{C} &\multicolumn{1}{c}{D} &\multicolumn{1}{c}{E} &\multicolumn{1}{c}{OK} &\multicolumn{1}{c}{attack} \\
\midrule
Q1 &0.01 &0.02 &0.23 &0.61 &0.79 &1.34 &0.24 \\
Q2 &0.00 &0.00 &0.01 &0.07 &0.08 &0.00 &0.05 \\
Q3 &0.01 &0.01 &0.22 &0.59 &0.88 &1.77 &0.23 \\
\bottomrule
& \multicolumn{2}{c}{\bf 8} & \multicolumn{5}{c}{\bf 9} \\
\cmidrule(lr){2-3}
\cmidrule(lr){4-8}
&\multicolumn{1}{c}{spam} &\multicolumn{1}{c}{vandalism} &\multicolumn{1}{c}{WP10B} &\multicolumn{1}{c}{WP10C} &\multicolumn{1}{c}{WP10FA} &\multicolumn{1}{c}{WP10GA} &\multicolumn{1}{c}{WP10Start} \\
\midrule
Q1 &1.17 &1.09 &2.08 &0.44 &0.16 &0.03 &1.12 \\
Q2 &0.03 &0.33 &1.71 &0.29 &0.15 &0.02 &0.06 \\
Q3 &1.26 &0.85 &2.63 &0.48 &0.24 &0.03 &1.19 \\
\bottomrule
&{\bf 9} &{\bf 10} &{\bf 11} &{\bf 12} &{\bf 13} &{\bf 15} &{\bf 16} \\
\midrule
&\multicolumn{1}{c}{WP10Stub} & & & & & & \\
\midrule
Q1 &3.03 &0.00 &21.75 &23.75 &0.00 &17.77 &15.37 \\
Q2 &1.03 &0.00 &14.43 &31.60 &37.89 &6.56 &10.70 \\
Q3 &2.68 &0.01 &21.05 &27.55 &24.15 &15.27 &28.93 \\
\bottomrule
&{\bf 17} &{\bf 18} & & & & & \\
\cmidrule{1-3}
Q1 &2.00 &3.19 & & & & & \\
Q2 &1.23 &1.17 & & & & & \\
Q3 &2.25 &3.45 & & & & & \\
\cmidrule[1pt]{1-3}
\end{tabular}
\end{table*}

Specifically, it shows the relative change in percentage between the original and synthetic data related to the first, second, and third quartiles of the synthetically generated samples. 
The minimal statistical variations observed in most features result from the synthetic data generation algorithm maintaining the inter-feature correlation.
The exceptions are the features that do not represent probabilistic values (\numrange{3}{5} and \numrange{11}{16}). After adding the \num{40000} synthetic to the original daily feature vectors, the final distribution of classes in the resulting balanced data set is \num{43357} revert and \num{41996} non-revert daily feature vectors.

\subsection{Stream processing}

Stream processing takes advantage of the insights obtained during the batch experiments to select the most promising features and classification algorithms. Mainly, the stream-based experiment explores the batch findings online. Moreover, graph and natural language descriptions of the model predictions are provided.

\subsubsection{Incremental profiling}

In the stream processing mode, the profiles of the editors are updated with each new daily feature vector (see Table \ref{tab:daily_features}), depending on the type of feature: 
\begin{itemize}
 \item Static features (1 and 2) remain unchanged since they correspond to identifiers;
 \item Average value features (\numrange{3}{18}) update their contents to the new incremental average; 
 \item Cumulative value features (19 and 20) update their contents to the new incremental sum.
\end{itemize}

\subsubsection{Incremental classification}

The following binary classification algorithms were selected from scikit-learn\footnote{Available at \url{https://scikit-learn.org/stable}, December 2023.} and scikit-multiflow\footnote{Available at \url{https://scikit-multiflow.readthedocs.io/en/stable/api/api.html}, December 2023.} and applied without hyper-parameter optimization.

\begin{itemize}
 \item \textsc{nb}\footnote{Available at \url{https://scikit-learn.org/stable/modules/naive_bayes.html}, December 2023.}
 \item \textsc{rc}\footnote{Available at \url{https://scikit-learn.org/stable/modules/generated/sklearn.linear_model.RidgeClassifier.html}, December 2023.}.
 \item \textsc{dt}\footnote{Available at \url{https://scikit-learn.org/stable/modules/generated/sklearn.tree.DecisionTreeClassifier.html}, December 2023.}
 \item \textsc{rf}\footnote{Available at \url{https://scikit-learn.org/stable/modules/tree.html} and \url{https://scikit-multiflow.readthedocs.io/en/stable/api/generated/skmultiflow.meta.AdaptiveRandomForestClassifier.html\#skmultiflow.meta.AdaptiveRandomForestClassifier}, December 2023.}
 \item \textsc{bc}\footnote{Available at \url{https://scikit-learn.org/stable/modules/generated/sklearn.ensemble.GradientBoostingClassifier.html}, December 2023.}
\end{itemize}

The following stream-based classification experiments were performed exclusively with the best classifier -- the \textsc{rf} model -- and best set of features -- set C. These experiments apply the \texttt{\small EvaluatePrequential}\footnote{Available at \url{https://scikit-multiflow.readthedocs.io/en/stable/api/generated/skmultiflow.evaluation.EvaluatePrequential.html}, December 2023.} algorithm that uses each incoming sample first to test and evaluate and, finally, to train the model.

The final classification experiments compare online and offline performance with the balanced data (\num{85353} samples) ordered chronologically. The online models are built from scratch and incrementally updated and evaluated, whereas the offline models are trained and then tested using distinct data partitions. The first set of experiments compares the classification results of the last \SI{90}{\percent} of the data for the online and the best offline model (trained with the first \SI{10}{\percent} of the data). The second set of experiments compares the results of the last \SI{10}{\percent} of the data for the online and the best offline model (trained with the first \SI{90}{\percent} of the data). Table \ref{tab:online_offline_balanced} displays these classification results. In both cases, the best offline results are obtained with set C features and decision tree classifiers (\textsc{dt} in the first case and \textsc{rf} in the second case). In the first set of experiments, the revert class (\#1) presents an online performance of near-\SI{90}{\percent}, \num{+69} percent points than the offline baseline in the recall metric, whereas, in the second set of experiments, the revert class (\#1) presents an online performance of near-\SI{100}{\percent}, \num{+3} percent points than the offline baseline in the recall metric. As expected, stream-based outperforms batch classification. 

\begin{figure}[htb]
\centering
\includegraphics[width=0.45\textwidth]{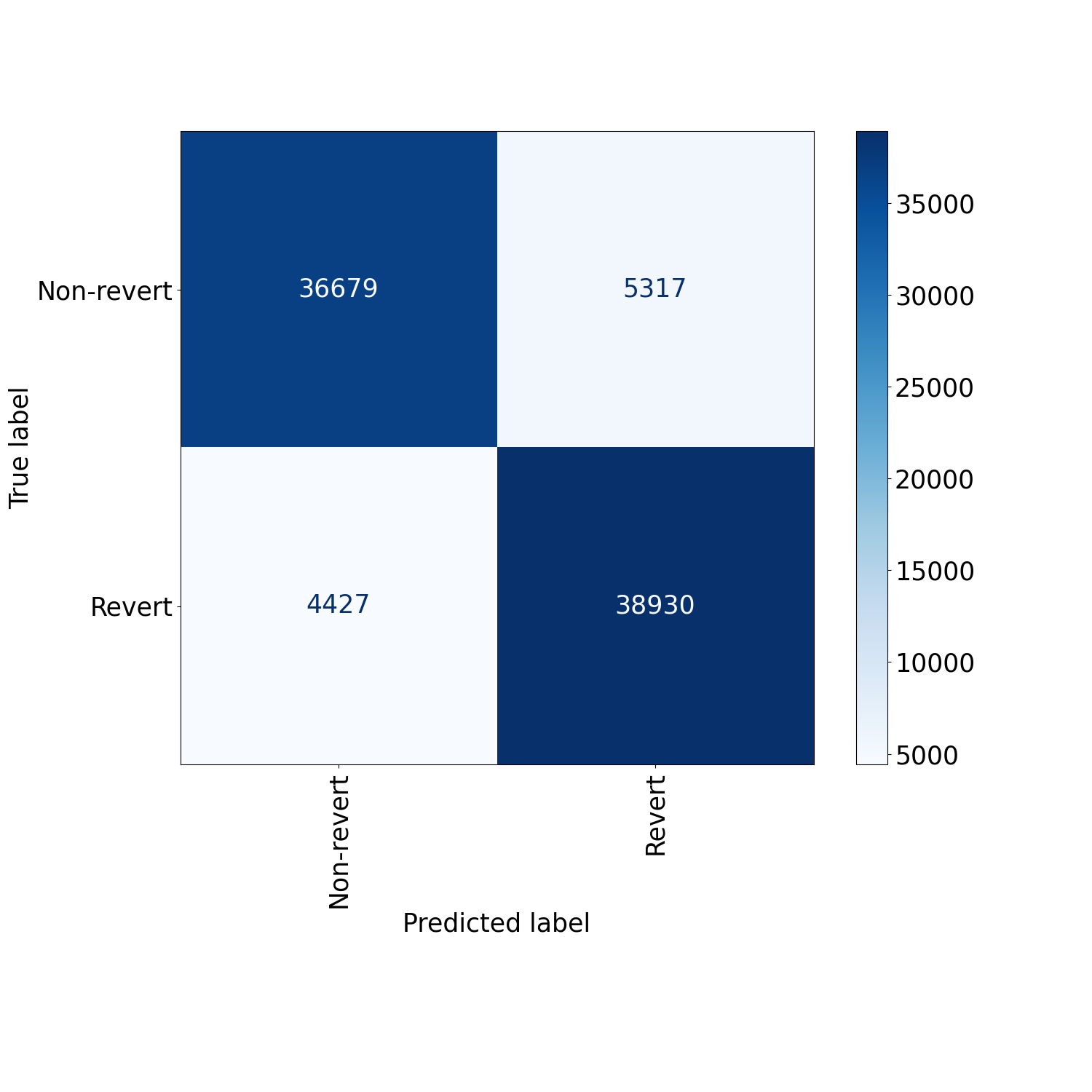}
\caption{\label{fig:confusion_matix} Confusion matrix of the online \textsc{rc} classifier with balanced data.}
\end{figure}

Figure \ref{fig:confusion_matix} displays the confusion matrix, showing the impact of false positives and negatives on the classification results\footnote{It corresponds to the online classification within the first set of experiments.}. Accordingly, the vast majority of the samples were correctly classified as they concentrated on the first diagonal of the matrix. Even though the model's performance is comparable in non-revert and revert detection tasks, the prediction error is slightly superior for the non-revert class. This means our solution is conservative when identifying editions to be reverted.

\begin{table*}[htb]
\centering
\scriptsize
\caption{\label{tab:online_offline_balanced}Online versus best offline revert classification results with balanced data.}
\begin{tabular}{lllcccccccccccS[table-format=4.2]}
    \toprule
    {\bf Processing} & {\bf Train/Update} & {\bf Test/Evaluate} & {\bf Classifier} & {\bf Accuracy} & \multicolumn{3}{c}{\bf Precision} & \multicolumn{3}{c}{\bf Recall} & \multicolumn{3}{c}{\bf \textit{F}-measure} & {\bf Time} \\
    \cmidrule(lr){6-8}
    \cmidrule(lr){9-11}
    \cmidrule(lr){12-14}
    & & & & & Macro & \#0 & \#1 & Macro & \#0 & \#1 & Macro & \#0 & \#1 & {(s)}\\
    \midrule
    Offline & First \SI{10}{\percent} & Last \SI{90}{\percent} & 
    \textsc{dt} & 0.48 & 0.53 & 0.45 & 0.60 & 0.52 & 0.83 & 0.21 & 0.45 & 0.59 & 0.31 & 1.96\\
    
    Online & \SI{100}{\percent} & Last \SI{90}{\percent} & \textsc{rf} & \bf 0.89 & \bf 0.89 & \bf 0.89 & \bf 0.88 & \bf 0.89 & \bf 0.87 & \bf 0.90 & \bf 0.89 & \bf 0.88 & \bf 0.89 & 1868.38 \\
    \midrule
    
    Offline & First \SI{90}{\percent} & Last \SI{10}{\percent} & \textsc{rf} & \bf 0.96 & 0.96 & 0.93 & \bf 0.99 & \bf 0.96 & \bf 0.99 & 0.93 & \bf 0.96 & \bf 0.96 & \bf 0.96 & 10.40 \\
    
    Online & \SI{100}{\percent} & Last \SI{10}{\percent} & \textsc{rf} & \bf 0.96 & \bf 0.97 & \bf 0.96 & 0.97 & \bf 0.96 & 0.97 & \bf 0.96 & \bf 0.96 & \bf 0.96 & \bf 0.96 & 1772.77 \\
    \bottomrule
\end{tabular}
\end{table*}

Finally, these results are better than those found in the literature, \textit{e.g.}, \num{+7}, \num{+16}, \num{+12} percent points in precision, recall, and \textit{F}-measure when compared with the closely related offline revert identification by \cite{Segall2013}. They explore an original data set from Simple English Wikipedia\footnote{Available at \url{https://simple.wikipedia.org/wiki/Main_Page}, December 2023.} composed of 3.1 million edits, \num{240000} articles and \num{175000} users. The experiments, performed with the Weka suite\footnote{Available at \url{https://www.cs.waikato.ac.nz/ml/weka}, December 2023.} and 10-fold cross-validation, use a subset of \num{825000} edits (\num{750000} for training and \num{75000} for test), ignoring article contents and relying on a Support Vector Machine classifier. Our comparable results, obtained with approximately one-tenth of the samples, are better than the proposed method.

\subsubsection{Explainability}
\label{sec:explainability}

The \textsc{rf} classifier provides the best results and builds and explores decision trees, which are interpretable models. These explanations cover the relevant subset of branches from root to classification leaf and materialize as the corresponding sub-graph and/or subset of learned rules.

Since the implemented online \textsc{rf} model uses ten estimators, the first step is to select the smallest decision path leading to the classification of a given sample. The next step uses the \texttt{\small get\_model\_description}\footnote{Available at \url{https://scikit-multiflow.readthedocs.io/en/stable/api/generated/skmultiflow.trees.HoeffdingTreeClassifier.html}, December 2023.} method from scikit-multiflow to traverse the selected decision tree and create the templates to display this knowledge in natural language. Listing \ref{lst:explicability_nl} provides four natural language explanations, two for each class (revert and non-revert), detailing the model decisions based on side and content-derived features as well as the predicted class (0 represents the non-revert class and 1, the revert class). The features correspond to Table \ref{tab:daily_features}.

\begin{figure}[htb]
\begin{lstlisting}[caption={Natural language explanations built from the \textsc{rf} classifier.},label={lst:explicability_nl}]
For sample 1, the model decision is based on the following facts:
 The average number of repeated links < 0.03
 Average ORES article quality probability - WP10GAAvg > 0.01
 Average ORES article quality probability - WP10FAAvg > 0.01
 Average ORES edit quality probability - damagingTrueAvg < 0.19
 Average ORES item quality probability - EAvg < 0.96
 Predicted class non-revert
For sample 2, the model decision is based on the following facts:
 Average ORES edit quality probability - damagingTrueAvg > 0.19
 Average ORES item quality probability - CAvg < 0.02
 The revision contains [`wiki']
 Predicted class revert
For sample 3, the model decision is based on the following facts:
 The average number of repeated links > 0.01
 Average ORES article quality probability - WP10BAvg < 0.22
 The revision contains [`speciality', `barbeque']
 Predicted class revert
For sample 4, the model decision is based on the following facts:
 The average number of repeated links < 0.01
 The revision contains [`jpg', `wikidata', `pgname', `long']
 Predicted class non-revert
\end{lstlisting}
\end{figure}

\begin{figure}[htb]
\centering
\includegraphics[width=0.45\textwidth]{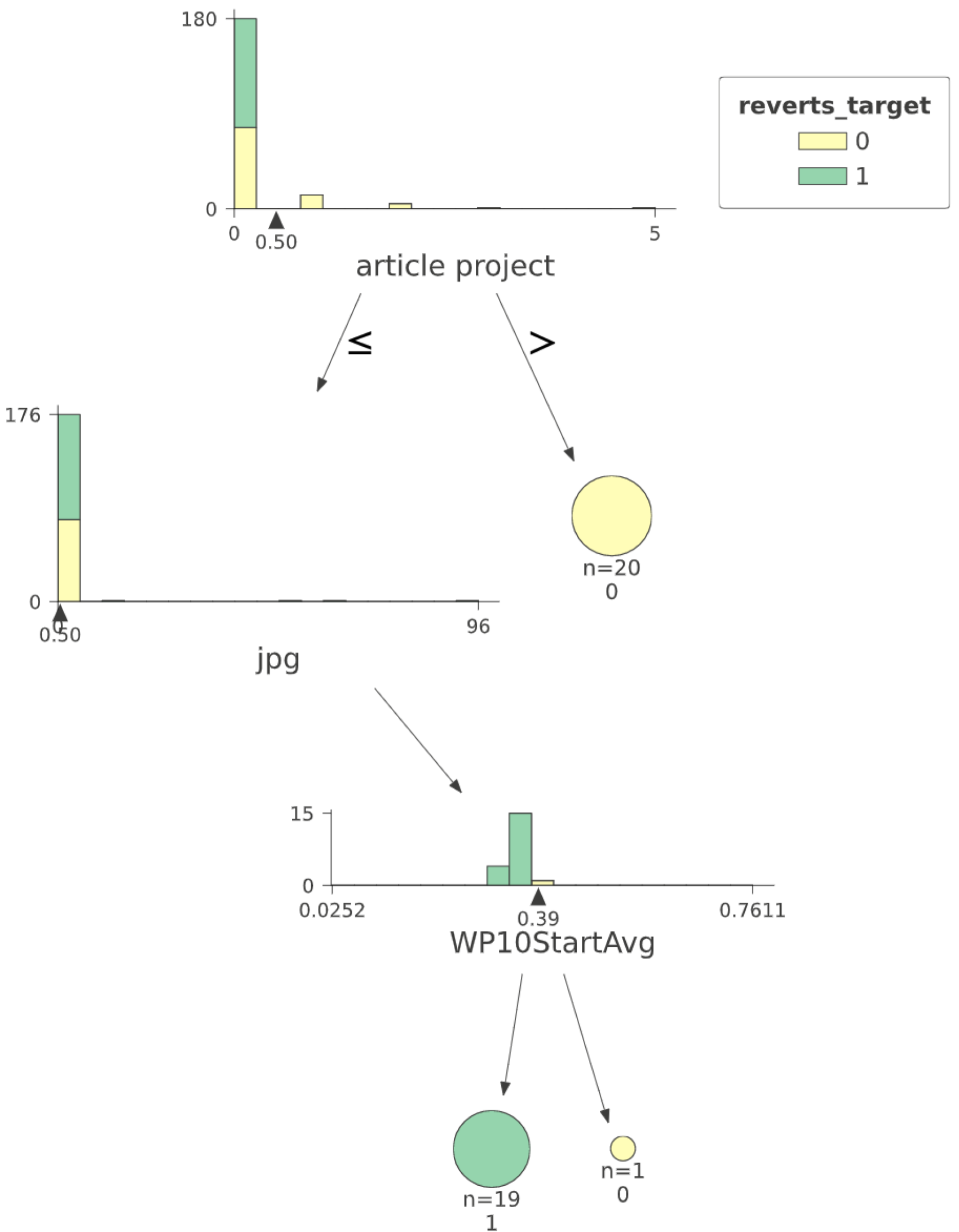}
\caption{\label{fig:partial_tree_enhanced} Enhanced graph-based explanations built from the \textsc{rf} classifier. The non-revert and revert classes correspond to 0 and 1, respectively.}
\end{figure}

\begin{figure}[htb]
\centering
\includegraphics[width=0.45\textwidth]{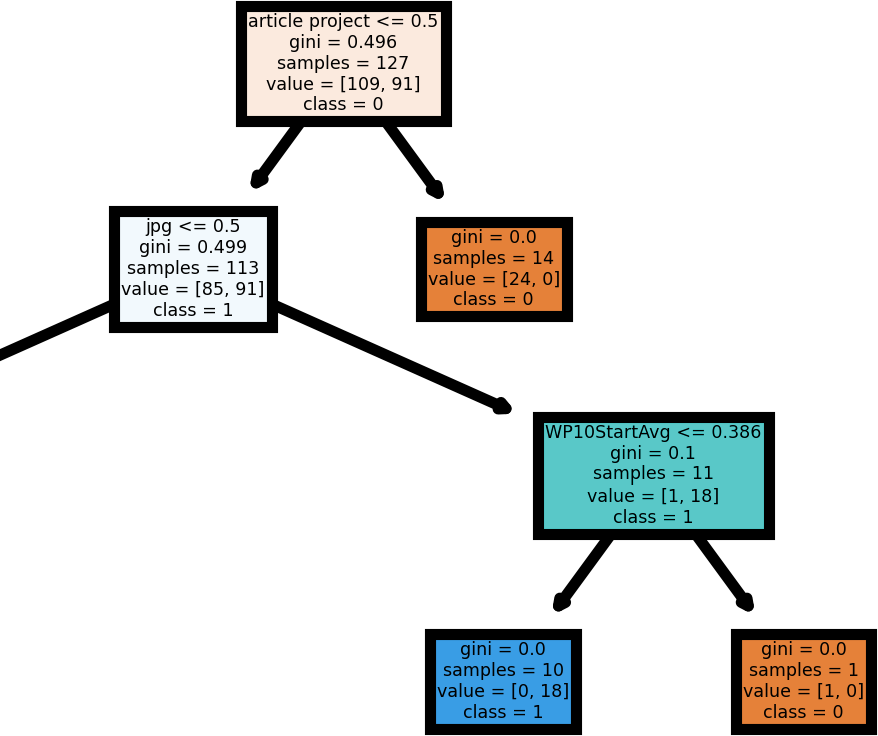}
\caption{\label{fig:partial_tree}Graph-based explanations built from the \textsc{rf} classifier. The non-revert and revert classes correspond to 0 and 1, respectively.}
\end{figure}

Figure \ref{fig:partial_tree_enhanced} displays a partial view of the decision tree learned and used by the \textsc{rf} algorithm in three cases. It depicts the revert and non-revert leaves using different colors (green for reverts and yellow for non-reverts) and styles and was obtained using \texttt{\small dtreeviz}\footnote{Available at \url{https://pypi.org/project/dtreeviz}, December 2023.} library. More in detail, the first decision is based on the bigram \texttt{\small article project} (feature 19 in Table \ref{tab:daily_features}). If the frequency of the bigram in the revision is superior to \num{0.5}, the reasoning continues through the right branch (predicted class non-revert); otherwise, it goes to the left. Using the \texttt{\small plot\_tree} library\footnote{Available at \url{https://scikit-learn.org/stable/modules/generated/sklearn.tree.plot_tree.html}, December 2023.} we can obtain the \texttt{\small gini} index (see Figure \ref{fig:partial_tree}). The latter coefficient expresses the probability of an incorrect prediction. For the first bifurcation, the \texttt{\small gini} index reported was close to 0.5. In the latter case, if the frequency of the unigram \texttt{\small jpg} is superior to \num{0.5}, the model checks if the average \textsc{ores} article quality probability (feature 9 in Table \ref{tab:daily_features} -- \texttt{\small WP10StartAvg}) is higher than \num{0.386}. If this test succeeds, the sample is classified as non-revert. Otherwise, it is considered a revert. As the decision tree is traversed, the gini index reduces, reaching 0.1 in the last bifurcation for this example.

Both explanations presented -- natural language and graph-based -- rely on the underlying model's interpretability and make the classifier's reasoning transparent and understandable for the user.

\section{Conclusions}
\label{sec:5}

Wiki platforms like Wikivoyage display articles created and maintained by a community of volunteer editors. This crowdsourcing model based on free content and open collaboration is vulnerable to unethical behavior, raising concerns about data quality in wiki repositories. One of the most critical issues behind this social manipulation is disinformation.

In light of the above, this work proposes a transparent, fair method to identify which wiki reviews to revert. Notably, the designed solution includes (\textit{i}) offline synthetic wiki data generation to ensure model fairness against class imbalance, (\textit{ii}) stream-based classification of wiki reviews supported by side and content-based features, and (\textit{iii}) interpretable \textsc{ml} models to explain to editors why their reviews were classified as reverts.

The stream-based experiments were performed with a balanced data set with \num{43357} revert and \num{41996} non-revert reviews made by \num{70260} editors to \num{3369} articles, resulting from the combination of collected and synthetic Wikivoyage data. The proposed method was evaluated using standard classification metrics and attained near-\SI{90}{\percent} in all classification metrics considered (precision, recall, \textit{F}-measure). This result shows that it is possible to explain and predict in real-time whether a review will be reverted and use this information to take preemptive actions to protect the quality of wiki articles and to help well-intentioned users improve their reviews.

The results obtained with the proposed method will immediately positively impact wiki users, who will enjoy more reliable content thanks to real-time review classification. The crowd volunteers will also benefit from explaining review classifications and reducing the editorial screening burden.

The challenges of integrating the proposed method into wiki platforms include (\textit{i}) labeling large volumes of data, (\textit{ii}) ensuring the privacy of sensitive data, and (\textit{iii}) assessing the trust of editors. These may be addressed by exploring \textsc{llm}s for data management, ad hoc algorithms for sensible data detection and removal, and reinforcement learning to model editor trust.

The future work plan will also explore the current result to automatically revert the identified reviews, outcast their authors, and evolve from the current mixed offline and online processing to a fully online processing pipeline, combined with hyper-parameter optimization, for further improvement.

\bibliography{bibliography.bib}{}

\begin{thebibliography}{10}
\providecommand{\url}[1]{#1}
\csname url@samestyle\endcsname
\providecommand{\newblock}{\relax}
\providecommand{\bibinfo}[2]{#2}
\providecommand{\BIBentrySTDinterwordspacing}{\spaceskip=0pt\relax}
\providecommand{\BIBentryALTinterwordstretchfactor}{4}
\providecommand{\BIBentryALTinterwordspacing}{\spaceskip=\fontdimen2\font plus
\BIBentryALTinterwordstretchfactor\fontdimen3\font minus \fontdimen4\font\relax}
\providecommand{\BIBforeignlanguage}[2]{{%
\expandafter\ifx\csname l@#1\endcsname\relax
\typeout{** WARNING: IEEEtran.bst: No hyphenation pattern has been}%
\typeout{** loaded for the language `#1'. Using the pattern for}%
\typeout{** the default language instead.}%
\else
\language=\csname l@#1\endcsname
\fi
#2}}
\providecommand{\BIBdecl}{\relax}
\BIBdecl

\bibitem{Alattar2021}
F.~Alattar and K.~Shaalan, ``{Using Artificial Intelligence to Understand What Causes Sentiment Changes on Social Media},'' \emph{IEEE Access}, vol.~9, pp. 61\,756--61\,767, 2021.

\bibitem{Jiang2023}
W.~Jiang and Y.~Sun, ``{Social-RippleNet: Jointly modeling of ripple net and social information for recommendation},'' \emph{Applied Intelligence}, vol.~53, pp. 3472--3487, 2023.

\bibitem{Zhang2021}
C.-B. Zhang, N.~Li, S.-H. Han, Y.-D. Zhang, and R.-J. Hou, ``{How to alleviate social loafing in online brand communities: The roles of community support and commitment},'' \emph{Electronic Commerce Research and Applications}, vol.~47, pp. 101\,051--101\,063, 2021.

\bibitem{Tajrian2023}
M.~Tajrian, A.~Rahman, M.~A. Kabir, and M.~R. Islam, ``{A Review of Methodologies for Fake News Analysis},'' \emph{IEEE Access}, vol.~11, pp. 73\,879--73\,893, 2023.

\bibitem{Lageard:2021}
V.~Lageard and C.~Paternotte, ``{Trolls, bans and reverts: simulating Wikipedia},'' \emph{Synthese}, vol. 198, pp. 451--470, 2021.

\bibitem{Mukherjee2021}
M.~Mukherjee and M.~Khushi, ``{SMOTE-ENC: A Novel SMOTE-Based Method to Generate Synthetic Data for Nominal and Continuous Features},'' \emph{Applied System Innovation}, vol.~4, no.~1, pp. 18--29, 2021.

\bibitem{Dankar2022}
F.~K. Dankar, M.~K. Ibrahim, and L.~Ismail, ``{A Multi-Dimensional Evaluation of Synthetic Data Generators},'' \emph{IEEE Access}, vol.~10, pp. 11\,147--11\,158, 2022.

\bibitem{lealwiki2020}
F.~Leal, B.~Veloso, B.~Malheiro, H.~González-Vélez, and J.~Carlos~Burguillo, ``{A 2020 Perspective on “Scalable Modelling and Recommendation Using Wiki-Based Crowdsourced Repositories”: Fairness, Scalability, and Real-Time Recommendation},'' \emph{Electronic Commerce Research and Applications}, vol.~40, pp. 100\,951--100\,952, 2020.

\bibitem{Heindorf:2019}
S.~Heindorf, Y.~Scholten, G.~Engels, and M.~Potthast, ``{Debiasing Vandalism Detection Models at Wikidata},'' in \emph{Proceedings of the The World Wide Web Conference}.\hskip 1em plus 0.5em minus 0.4em\relax Association for Computing Machinery, 2019, p. 670–680.

\bibitem{Harpalani:2011}
M.~Harpalani, M.~Hart, S.~Singh, R.~Johnson, and Y.~Choi, ``{Language of Vandalism: Improving Wikipedia Vandalism Detection via Stylometric Analysis},'' in \emph{Proceedings of the Annual Meeting of the Association for Computational Linguistics: Human Language Technologies}.\hskip 1em plus 0.5em minus 0.4em\relax Association for Computational Linguistics, 2011, p. 83–88.

\bibitem{Dauber:2017}
E.~Dauber, R.~Overdorf, and R.~Greenstadt, ``{Stylometric Authorship Attribution of Collaborative Documents},'' in \emph{Proceedings of the Cyber Security Cryptography and Machine Learning Conference}.\hskip 1em plus 0.5em minus 0.4em\relax Springer, 2017, pp. 115--135.

\bibitem{Martinez-Rico:2019}
J.~R. Martinez-Rico, J.~Martinez-Romo, and L.~Araujo, ``{Can deep learning techniques improve classification performance of vandalism detection in Wikipedia?}'' \emph{Engineering Applications of Artificial Intelligence}, vol.~78, pp. 248--259, 2019.

\bibitem{Zhao:2011}
H.~Zhao, W.~Kallander, T.~Gbedema, H.~Johnson, and F.~Wu, ``{Read What You Trust: An Open Wiki Model Enhanced by Social Context},'' in \emph{Proceedings of the IEEE International Conference on Privacy, Security, Risk and Trust and IEEE Third International Conference on Social Computing}, 2011, pp. 370--379.

\bibitem{Zhao:2013}
H.~Zhao, W.~Kallander, H.~Johnson, and S.~F. Wu, ``{SmartWiki: A reliable and conflict-refrained Wiki model based on reader differentiation and social context analysis},'' \emph{Knowledge-Based Systems}, vol.~47, pp. 53--64, 2013.

\bibitem{Adler:2012}
B.~T. Adler, ``{WikiTrust: Content-Driven Reputation for the Wikipedia},'' Ph.D. dissertation, University of California, 2012.

\bibitem{Kardan:2015}
A.~A. Kardan, R.~Salarmehr, and A.~Farshad, ``{SigmoRep: A Robust Reputation Model for Open Collaborative Environments},'' in \emph{Proceedings of the IEEE International Conference on Trust, Security and Privacy in Computing and Communications}, 2014, pp. 505--510.

\bibitem{Paul:2015}
P.~P. Paul, M.~Sultana, S.~A. Matei, and M.~Gavrilova, ``{Editing Behavior to Recognize Authors of Crowdsourced Content},'' in \emph{Proceedings of the IEEE International Conference on Systems, Man, and Cybernetics}, 2015, pp. 1676--1681.

\bibitem{lealwiki2019}
F.~Leal, B.~M. Veloso, B.~Malheiro, H.~González-Vélez, and J.~C. Burguillo, ``{Scalable modelling and recommendation using wiki-based crowdsourced repositories},'' \emph{Electronic Commerce Research and Applications}, vol.~33, pp. 100\,817--100\,825, 2019.

\bibitem{Garcia-Mendez2022}
S.~García-Méndez, F.~Leal, B.~Malheiro, J.~C. Burguillo-Rial, B.~Veloso, A.~E. Chis, and H.~González–Vélez, ``{Simulation, modelling and classification of wiki contributors: Spotting the good, the bad, and the ugly},'' \emph{Simulation Modelling Practice and Theory}, vol. 120, pp. 102\,616--102\,628, 2022.

\bibitem{Joshi:2020}
N.~Joshi, F.~Spezzano, M.~Green, and E.~Hill, ``{Detecting Undisclosed Paid Editing in Wikipedia},'' in \emph{Proceedings of The Web Conference}.\hskip 1em plus 0.5em minus 0.4em\relax Association for Computing Machinery, 2020, pp. 2899--2905.

\bibitem{Bertsch:2021}
A.~Bertsch and S.~Bethard, ``{Detection of Puffery on the {E}nglish {W}ikipedia},'' in \emph{Proceedings of the Workshop on Noisy User-generated Text}.\hskip 1em plus 0.5em minus 0.4em\relax Association for Computational Linguistics, 2021, pp. 329--333.

\bibitem{Flock:2012}
F.~Fl\"{o}ck, D.~Vrande\v{c}i\'{c}, and E.~Simperl, ``{Revisiting Reverts: Accurate Revert Detection in Wikipedia},'' in \emph{Proceedings of the ACM Conference on Hypertext and Social Media}.\hskip 1em plus 0.5em minus 0.4em\relax Association for Computing Machinery, 2012, p. 3–12.

\bibitem{Segall2013}
J.~Segall and R.~Greenstadt, ``{The Illiterate Editor: Metadata-Driven Revert Detection in Wikipedia},'' in \emph{Proceedings of the International Symposium on Open Collaboration}.\hskip 1em plus 0.5em minus 0.4em\relax Association for Computing Machinery, 2013, pp. 1--8.

\bibitem{Kiesel:2017}
J.~Kiesel, M.~Potthast, M.~Hagen, and B.~Stein, ``{Spatio-Temporal Analysis of Reverted Wikipedia Edits},'' in \emph{Proceedings of the International AAAI Conference on Web and Social Media}, vol.~11, 2017, pp. 122--131.

\bibitem{Ibrahim:2018}
M.~Ibrahim, M.~Torki, and N.~El-Makky, ``{Imbalanced Toxic Comments Classification Using Data Augmentation and Deep Learning},'' in \emph{Proceedings of the IEEE International Conference on Machine Learning and Applications}, 2018, pp. 875--878.

\bibitem{Chakrabarty:2020}
N.~Chakrabarty, ``{A Machine Learning Approach to Comment Toxicity Classification},'' in \emph{Proceedings of the Computational Intelligence in Pattern Recognition Conference}.\hskip 1em plus 0.5em minus 0.4em\relax Springer, 2020, pp. 183--193.

\bibitem{Potthast:2008}
M.~Potthast, B.~Stein, and R.~Gerling, ``{Automatic Vandalism Detection in Wikipedia},'' in \emph{Proceedings of Advances in Information Retrieval Conference}.\hskip 1em plus 0.5em minus 0.4em\relax Springer, 2008, pp. 663--668.

\bibitem{Adler:2011}
B.~T. Adler, L.~de~Alfaro, S.~M. Mola-Velasco, P.~Rosso, and A.~G. West, ``{Wikipedia Vandalism Detection: Combining Natural Language, Metadata, and Reputation Features},'' in \emph{Proceedings of the Computational Linguistics and Intelligent Text Processing Conference}.\hskip 1em plus 0.5em minus 0.4em\relax Springer, 2011, pp. 277--288.

\bibitem{Javanmardi:2011}
S.~Javanmardi, D.~W. McDonald, and C.~V. Lopes, ``{Vandalism Detection in Wikipedia: A High-Performing, Feature-Rich Model and Its Reduction through Lasso},'' in \emph{Proceedings of the International Symposium on Wikis and Open Collaboration}.\hskip 1em plus 0.5em minus 0.4em\relax Association for Computing Machinery, 2011, p. 82–90.

\bibitem{Mola-Velasco:2011}
S.~M. Mola-Velasco, ``{Wikipedia Vandalism Detection},'' in \emph{Proceedings of the International Conference Companion on World Wide Web}.\hskip 1em plus 0.5em minus 0.4em\relax Association for Computing Machinery, 2011, p. 391–396.

\bibitem{Alfonseca:2013}
E.~Alfonseca, G.~Garrido, J.~Y. Delort, and A.~Peñas, ``{WHAD: Wikipedia historical attributes data},'' \emph{Language Resources and Evaluation}, vol.~47, pp. 1163--1190, 2013.

\bibitem{Tran:2013}
K.-N. Tran and P.~Christen, ``{Cross Language Prediction of Vandalism on Wikipedia Using Article Views and Revisions},'' in \emph{Proceedings of the Advances in Knowledge Discovery and Data Mining Conference}.\hskip 1em plus 0.5em minus 0.4em\relax Springer, 2013, pp. 268--279.

\bibitem{Kumar:2015}
S.~Kumar, F.~Spezzano, and V.~Subrahmanian, ``{VEWS: A Wikipedia Vandal Early Warning System},'' in \emph{Proceedings of the ACM SIGKDD International Conference on Knowledge Discovery and Data Mining}.\hskip 1em plus 0.5em minus 0.4em\relax Association for Computing Machinery, 2015, p. 607–616.

\bibitem{Heindorf:2016}
S.~Heindorf, M.~Potthast, B.~Stein, and G.~Engels, ``{Vandalism Detection in Wikidata},'' in \emph{Proceedings of the ACM International on Conference on Information and Knowledge Management}.\hskip 1em plus 0.5em minus 0.4em\relax Association for Computing Machinery, 2016, p. 327–336.

\bibitem{Shulhan:2016}
M.~Shulhan and D.~H. Widyantoro, ``{Detecting vandalism on English Wikipedia using LNSMOTE resampling and Cascaded Random Forest classifier},'' in \emph{Proceedings of the International Conference On Advanced Informatics: Concepts, Theory And Application}, 2016, pp. 1--6.

\bibitem{Heindorf:2017}
S.~Heindorf, M.~Potthast, G.~Engels, and B.~Stein, ``{Overview of the Wikidata Vandalism Detection Task at WSDM Cup 2017},'' in \emph{Proceedings of the Web Search and Data Mining Cup}, 2017, pp. 1--9.

\bibitem{Sarabadani:2017}
A.~Sarabadani, A.~Halfaker, and D.~Taraborelli, ``{Building Automated Vandalism Detection Tools for Wikidata},'' in \emph{Proceedings of the International Conference on World Wide Web Companion}.\hskip 1em plus 0.5em minus 0.4em\relax International World Wide Web Conferences Steering Committee, 2017, p. 1647–1654.

\bibitem{Liu:2019}
Z.~Liu and A.~Lu, ``{Explainable Visualization for Interactive Exploration of CNN on Wikipedia Vandal Detection},'' in \emph{Proceedings of the IEEE International Conference on Big Data}, 2019, pp. 2354--2363.

\bibitem{Sarkar:2019}
S.~Sarkar, B.~P. Reddy, S.~Sikdar, and A.~Mukherjee, ``{StRE: Self attentive edit quality prediction in Wikipedia},'' in \emph{Proceedings of the Annual Meeting of the Association for Computational Linguistics}.\hskip 1em plus 0.5em minus 0.4em\relax Association for Computational Linguistics, 2019, pp. 3962--3972.

\bibitem{Asthana:2021}
S.~Asthana, S.~Tobar~Thommel, A.~L. Halfaker, and N.~Banovic, ``{Automatically Labeling Low Quality Content on Wikipedia By Leveraging Patterns in Editing Behaviors},'' in \emph{Proceedings of the ACM on Human-Computer Interaction Conference}, vol.~5.\hskip 1em plus 0.5em minus 0.4em\relax Association for Computing Machinery, 2021, pp. 1--23.

\bibitem{Wong:2021}
K.~Wong, M.~Redi, and D.~Saez-Trumper, ``{Wiki-Reliability: A Large Scale Dataset for Content Reliability on Wikipedia},'' in \emph{Proceedings of the International ACM SIGIR Conference on Research and Development in Information Retrieval}.\hskip 1em plus 0.5em minus 0.4em\relax Association for Computing Machinery, 2021, p. 2437–2442.

\bibitem{Ruprechter:2020}
T.~Ruprechter, T.~Santos, and D.~Helic, ``{Relating Wikipedia article quality to edit behavior and link structure},'' \emph{Applied Network Science}, vol.~5, no.~1, pp. 61--80, 2020.

\bibitem{dang2016measuring}
Q.-V. Dang and C.-L. Ignat, ``{Measuring Quality of Collaboratively Edited Documents: The Case of Wikipedia},'' in \emph{Proceedings of the IEEE International Conference on Collaboration and Internet Computing}, 2016, pp. 266--275.

\bibitem{dang2017end}
Q.~V. Dang and C.~L. Ignat, ``{An End-to-End Learning Solution for Assessing the Quality of Wikipedia Articles},'' in \emph{Proceedings of the International Symposium on Open Collaboration}.\hskip 1em plus 0.5em minus 0.4em\relax Association for Computing Machinery, 2017, pp. 1--10.

\bibitem{lewoniewski2017relative}
W.~Lewoniewski and K.~W{\k{e}}cel, ``{Relative Quality Assessment of Wikipedia Articles in Different Languages Using Synthetic Measure},'' in \emph{Proceedings of the Business Information Systems Workshops}.\hskip 1em plus 0.5em minus 0.4em\relax Springer, 2017, pp. 282--292.

\bibitem{Alkharashi:2018}
A.~Alkharashi and J.~Jose, ``{Vandalism on Collaborative Web Communities: An Exploration of Editorial Behaviour in Wikipedia},'' in \emph{Proceedings of the Spanish Conference on Information Retrieval}.\hskip 1em plus 0.5em minus 0.4em\relax Association for Computing Machinery, 2018, pp. 1--4.

\bibitem{Parmar2019}
A.~Parmar, R.~Katariya, and V.~Patel, ``{A Review on Random Forest: An Ensemble Classifier},'' in \emph{Proceedings of the International Conference on Intelligent Data Communication Technologies and Internet of Things}.\hskip 1em plus 0.5em minus 0.4em\relax Springer, 2019, pp. 758--763.

\bibitem{Halfaker:2020}
A.~Halfaker and R.~S. Geiger, ``{ORES: Lowering Barriers with Participatory Machine Learning in Wikipedia},'' in \emph{Proceedings of the ACM on Human-Computer Interaction Conference}, vol.~4.\hskip 1em plus 0.5em minus 0.4em\relax Association for Computing Machinery, 2020, pp. 1--37.

\bibitem{Frey2023}
B.~B. Frey, \emph{{Multiple Logistic Regression}}.\hskip 1em plus 0.5em minus 0.4em\relax SAGE Publications, Inc., 2023.

\bibitem{Mistry2021}
M.~Mistry, D.~Letsios, G.~Krennrich, R.~M. Lee, and R.~Misener, ``{Mixed-Integer Convex Nonlinear Optimization with Gradient-Boosted Trees Embedded},'' \emph{INFORMS Journal on Computing}, vol.~33, no.~3, pp. 1103--1119, 2021.

\bibitem{Adadi2018}
A.~Adadi and M.~Berrada, ``{Peeking Inside the Black-Box: A Survey on Explainable Artificial Intelligence (XAI)},'' \emph{IEEE Access}, vol.~6, pp. 52\,138--52\,160, 2018.

\bibitem{Mollas2023}
I.~Mollas, N.~Bassiliades, and G.~Tsoumakas, ``{LioNets: a neural-specific local interpretation technique exploiting penultimate layer information},'' \emph{Applied Intelligence}, vol.~53, pp. 2538--2563, 2023.

\bibitem{Subramanian:2019}
S.~S. Subramanian, P.~Pushparaj, Z.~Liu, and A.~Lu, ``{Explainable Visualization of Collaborative Vandal Behaviors in Wikipedia},'' in \emph{Proceedings of the IEEE Symposium on Visualization for Cyber Security}, 2019, pp. 1--5.

\bibitem{Mahajan:2021}
A.~Mahajan, D.~Shah, and G.~Jafar, ``{Explainable AI Approach Towards Toxic Comment Classification},'' in \emph{Proceedings of the Emerging Technologies in Data Mining and Information Security Conference}.\hskip 1em plus 0.5em minus 0.4em\relax Springer, 2021, pp. 849--858.

\bibitem{Sarker:2020}
M.~K. Sarker, J.~Schwartz, P.~Hitzler, L.~Zhou, S.~Nadella, B.~Minnery, I.~Juvina, M.~L. Raymer, and W.~R. Aue, ``{Wikipedia Knowledge Graph for Explainable {AI}},'' in \emph{Proceedings of the Knowledge Graphs and Semantic Web Conference}.\hskip 1em plus 0.5em minus 0.4em\relax Springer, 2020, pp. 72--87.

\bibitem{Klein:2021}
N.~Klein, F.~Ilievski, and P.~Szekely, ``{Generating Explainable Abstractions for Wikidata Entities},'' in \emph{Proceedings of the on Knowledge Capture Conference}.\hskip 1em plus 0.5em minus 0.4em\relax Association for Computing Machinery, 2021, p. 89–96.

\bibitem{Ribeiro:2016}
M.~T. Ribeiro, S.~Singh, and C.~Guestrin, ``{``Why Should I Trust You?'': Explaining the Predictions of Any Classifier},'' in \emph{Proceedings of the ACM SIGKDD International Conference on Knowledge Discovery and Data Mining}.\hskip 1em plus 0.5em minus 0.4em\relax Association for Computing Machinery, 2016, p. 1135–1144.

\bibitem{Lundberg:2017}
S.~M. Lundberg and S.-I. Lee, ``{A Unified Approach to Interpreting Model Predictions},'' in \emph{Proceedings of the International Conference on Neural Information Processing Systems}.\hskip 1em plus 0.5em minus 0.4em\relax Curran Associates Inc., 2017, p. 4768–4777.

\bibitem{Leal:2022}
F.~Leal, S.~García-Méndez, B.~Malheiro, and J.~C. Burguillo, ``{Explanation Plug-In for Stream-Based Collaborative Filtering},'' in \emph{Lecture Notes in Networks and Systems}, vol. 468 LNNS, 2022, pp. 42--51.

\bibitem{Risch:2020}
J.~Risch, R.~Ruff, and R.~Krestel, ``{Offensive Language Detection Explained},'' in \emph{Proceedings of the Workshop on Trolling, Aggression and Cyberbullying}, vol.~34, 2020, pp. 29--47.

\bibitem{Qureshi:2019}
M.~A. Qureshi and D.~Greene, ``{EVE: Explainable Vector Based Embedding Technique Using Wikipedia},'' \emph{Journal of Intelligent Information Systems}, vol.~53, no.~1, p. 137–165, 2019.

\bibitem{Ye:2021}
Z.~Ye, X.~Yuan, S.~Gaur, A.~Halfaker, J.~Forlizzi, and H.~Zhu, ``{Wikipedia ORES Explorer: Visualizing Trade-Offs For Designing Applications With Machine Learning API},'' in \emph{Proceedings of the Designing Interactive Systems Conference}.\hskip 1em plus 0.5em minus 0.4em\relax Association for Computing Machinery, 2021, p. 1554–1565.

\bibitem{Lewandowski:2011}
D.~Lewandowski and U.~Spree, ``{Ranking of Wikipedia articles in search engines revisited: Fair ranking for reasonable quality?}'' \emph{Journal of the American Society for Information Science and Technology}, vol.~62, no.~1, pp. 117--132, 2011.

\bibitem{Ross:2014}
S.~Ross, ``{Your Day in `Wiki-Court': ADR, Fairness, and Justice in Wikipedia's Global Community},'' \emph{Osgoode Legal Studies Research Paper}, vol.~10, pp. 1--21, 2014.

\bibitem{Tramullas:2016}
J.~Tramullas, P.~Garrido-Picazo, and A.~I. Sánchez-Casabón, ``{Research on Wikipedia Vandalism: A Brief Literature Review},'' in \emph{Proceedings of the Spanish Conference on Information Retrieval}.\hskip 1em plus 0.5em minus 0.4em\relax Association for Computing Machinery, 2016, pp. 1--4.

\bibitem{DeWinter2016}
J.~C.~F. De~Winter, S.~D. Gosling, and J.~Potter, ``{Comparing the Pearson and Spearman correlation coefficients across distributions and sample sizes: A tutorial using simulations and empirical data.}'' \emph{Psychological Methods}, vol.~21, no.~3, pp. 273--290, 2016.

\bibitem{Mi2020}
J.-X. Mi, A.-D. Li, and L.-F. Zhou, ``{Review Study of Interpretation Methods for Future Interpretable Machine Learning},'' \emph{IEEE Access}, vol.~8, pp. 191\,969--191\,985, 2020.

\bibitem{Haffar2022}
R.~Haffar, D.~Sánchez, and J.~Domingo-Ferrer, ``{Explaining predictions and attacks in federated learning via random forests},'' \emph{Applied Intelligence}, vol.~53, no.~1, pp. 169--185, 2022.

\bibitem{Hossin2023}
M.~M. Hossin, F.~M. J.~M. Shamrat, M.~R. Bhuiyan, R.~A. Hira, T.~Khan, and S.~Molla, ``{Breast cancer detection: an effective comparison of different machine learning algorithms on the Wisconsin dataset},'' \emph{Bulletin of Electrical Engineering and Informatics}, vol.~12, pp. 2446--2456, 2023.

\bibitem{Berrar2019NB}
D.~Berrar, ``{Bayes' Theorem and Naive Bayes Classifier},'' in \emph{Encyclopedia of Bioinformatics and Computational Biology}.\hskip 1em plus 0.5em minus 0.4em\relax Elsevier, 2019, vol. 1-3, pp. 403--412.

\bibitem{sarker2021machine}
I.~H. Sarker, ``{Machine learning: Algorithms, real-world applications and research directions},'' \emph{SN computer science}, vol.~2, no.~3, pp. 160--180, 2021.

\bibitem{Trabelsi2019}
A.~Trabelsi, Z.~Elouedi, and E.~Lefevre, ``{Decision tree classifiers for evidential attribute values and class labels},'' \emph{Fuzzy Sets and Systems}, vol. 366, pp. 46--62, 2019.

\bibitem{Bentejac2021}
C.~Bent{\'{e}}jac, A.~Cs{\"{o}}rgő, and G.~Mart{\'{i}}nez-Mu{\~{n}}oz, ``{A comparative analysis of gradient boosting algorithms},'' \emph{Artificial Intelligence Review}, vol.~54, no.~3, pp. 1937--1967, 2021.

\bibitem{Okkalioglu2022}
M.~Okkalioglu and B.~D. Okkalioglu, ``{AFE-MERT: imbalanced text classification with abstract feature extraction},'' \emph{Applied Intelligence}, vol.~52, no.~9, pp. 10\,352--10\,368, 2022.

\bibitem{Hafeez2023}
A.~Hafeez, T.~Ali, A.~Nawaz, S.~U. Rehman, A.~I. Mudasir, A.~A. Alsulami, and A.~Alqahtani, ``{Addressing Imbalance Problem for Multi Label Classification of Scholarly Articles},'' \emph{IEEE Access}, vol.~11, pp. 74\,500--74\,516, 2023.

\bibitem{Vanacore2023}
A.~Vanacore, M.~S. Pellegrino, and A.~Ciardiello, ``{Evaluating classifier predictive performance in multi‐class problems with balanced and imbalanced data sets},'' \emph{Quality and Reliability Engineering International}, vol.~39, pp. 651--669, 2023.

\bibitem{Berrar2019}
D.~Berrar, ``{Cross-Validation},'' in \emph{Encyclopedia of Bioinformatics and Computational Biology}.\hskip 1em plus 0.5em minus 0.4em\relax Academic Press, 2019, pp. 542--545.

\end{thebibliography}
\bibliographystyle{IEEEtran}

\begin{IEEEbiography}[{\includegraphics[width=1in,height=1.25in,clip,keepaspectratio]{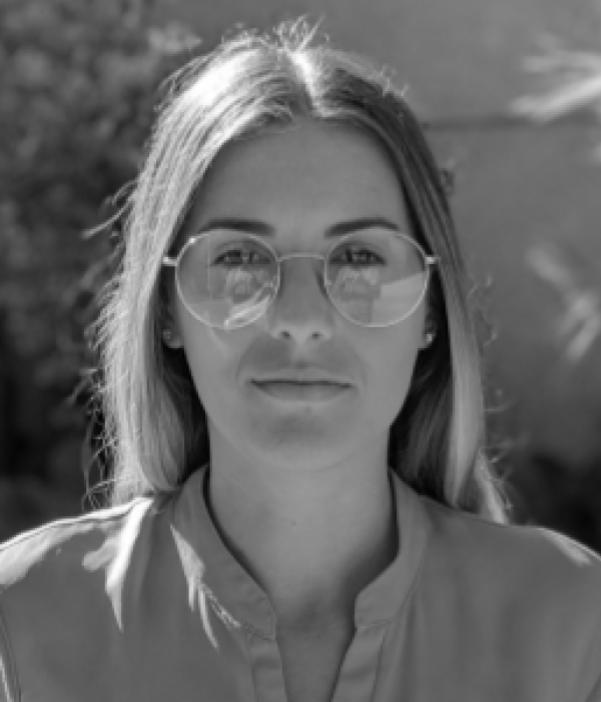}}]{Silvia García-Méndez} received a Ph.D. in Information and Communication Technologies from the University of Vigo, Spain, in 2021. Since 2015, she has worked as a researcher with the Information Technologies Group at the University of Vigo. She is collaborating with foreign research centers as part of her postdoctoral stage. Her research interests include Natural Language Processing techniques and Machine Learning algorithms.
\end{IEEEbiography}

\begin{IEEEbiography}[{\includegraphics[width=1in,height=1.25in,clip,keepaspectratio]{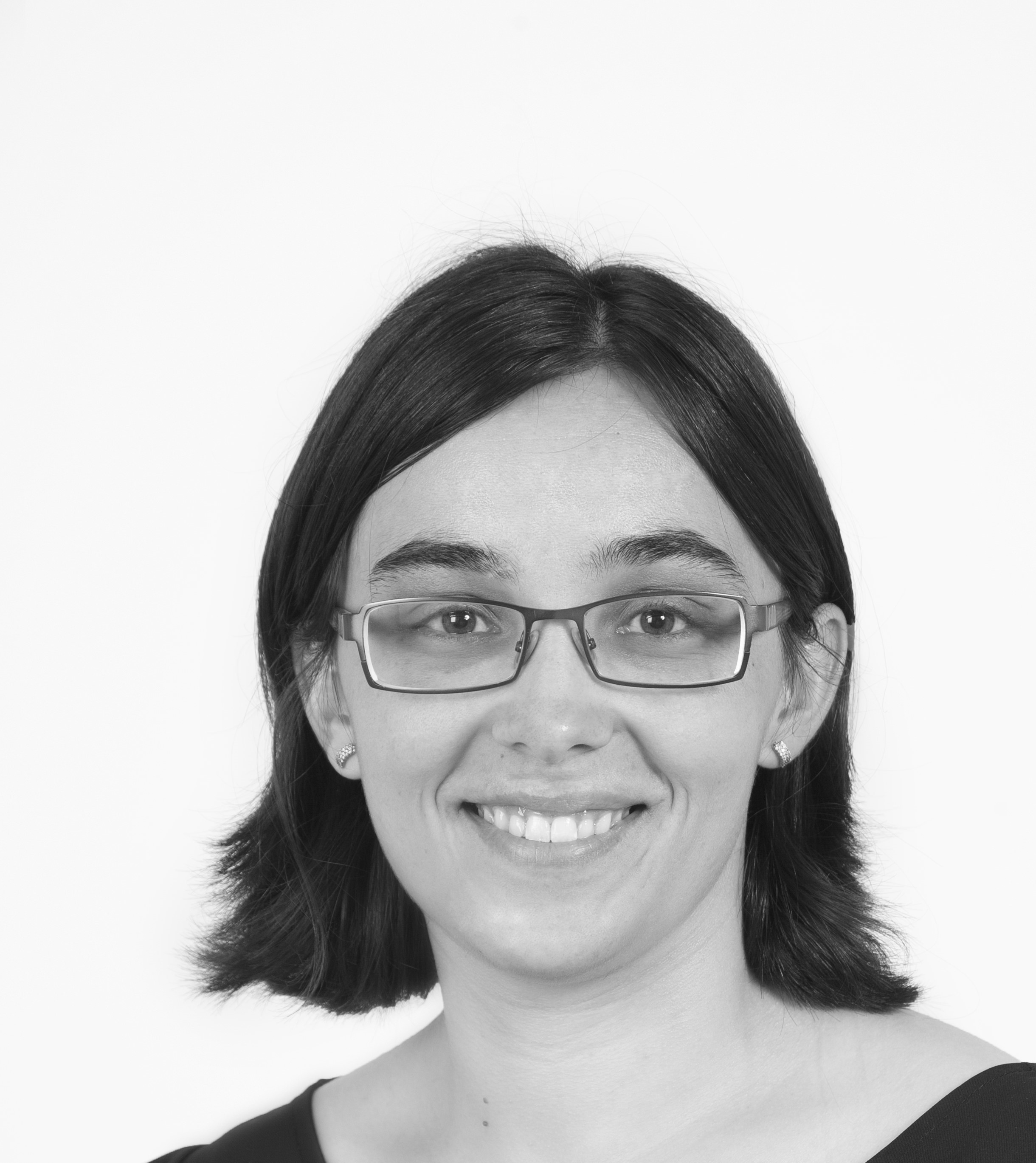}}]{Fátima Leal} holds a Ph.D. in Information and Communication Technologies from the University of Vigo, Spain. She is an Assistant Professor at Universidade Portucalense, Portugal, and a researcher at REMIT (Research on Economics, Management, and Information Technologies). Following a full-time postdoctoral fellowship funded by the European Commission, she continues collaborating with The Cloud Competency Center at the National College of Ireland in Dublin. Her research is based on crowdsourced information, including trust and reputation, Big Data, Data Streams, and Recommendation Systems. Recently, she has been exploring blockchain technologies for responsible data processing.
\end{IEEEbiography}

\begin{IEEEbiography}[{\includegraphics[width=1in,height=1.25in,clip,keepaspectratio]{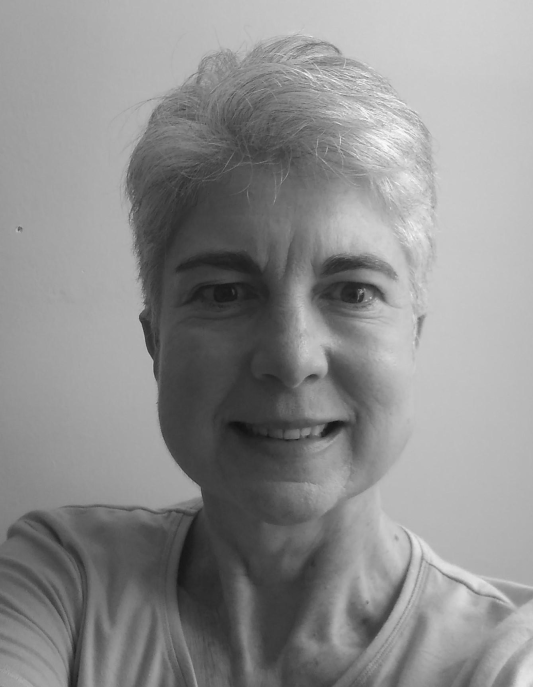}}]{Benedita Malheiro} is a Coordinator Professor at Instituto Superior de Engenharia do Porto, the School of Engineering of the Polytechnic of Porto, and senior researcher at \textsc{inesc} \textsc{tec}, Portugal. She holds a Ph.D. and an M.Sc. in Electrical Engineering and Computers and a five-year graduation in Electrical Engineering from the University of Porto. Her research interests include Artificial Intelligence, Computer Science, and Engineering Education. She is a member of the Association for the Advancement of Artificial Intelligence (\textsc{aaai}), the Portuguese Association for Artificial Intelligence (\textsc{appia}), the Association for Computing Machinery (\textsc{acm}), and the Professional Association of Portuguese Engineers (\textsc{oe}).
\end{IEEEbiography}

\begin{IEEEbiography}[{\includegraphics[width=1in,height=1.25in,clip,keepaspectratio]{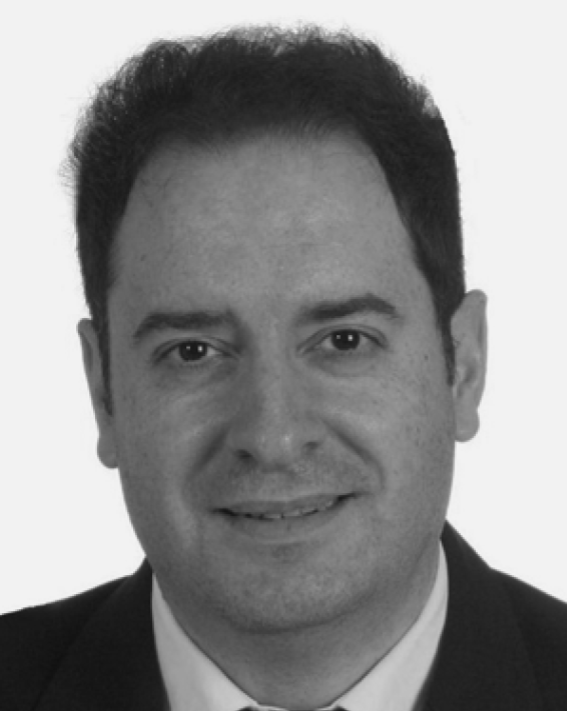}}]{Juan Carlos Burguillo-Rial} received the Ph.D. degree in Telematics in 2001 at the University of Vigo, Spain. He is currently a Full Professor at the Department of Telematic Engineering and a researcher at the AtlanTTic Research Center in Telecommunication Technologies at the University of Vigo. He has directed and participated in several R\&D projects in the areas of Telematics and Computer Science in national and international calls. He is a regular reviewer of several international conferences and journals including the Journal of Autonomous Agents and Multi-agent Systems, Computers and Education, Engineering Applications of Artificial Intelligence, Computers and Mathematics with Applications, and the Journal of Network and Computer Applications among others. He is also the area editor of the journal Simulation Modelling Practice and Theory (SIMPAT) in his topics of interest: intelligent systems, evolutionary game theory, self-organization, and complex adaptive systems.
\end{IEEEbiography}

\EOD

\end{document}